\documentclass{article}

\usepackage{hyperref}
\usepackage{siunitx}
\usepackage[accepted]{icml2023}
\usepackage{soul}

\usepackage{amsmath}
\usepackage{amssymb}
\usepackage{mathtools}
\usepackage{amsthm}
\usepackage{amsfonts}       % blackboard math symbols

\usepackage[capitalize,noabbrev]{cleveref}

\usepackage{microtype}
\usepackage{graphicx}
\usepackage{overpic}
\usepackage{booktabs} % for professional tables
\usepackage{multirow}

\usepackage{lipsum}

\theoremstyle{plain}

\theoremstyle{definition}

\theoremstyle{remark}

\usepackage[textsize=tiny]{todonotes}

\icmltitlerunning{Sample and Predict Your Latent: Modality-free Sequential Disentanglement via Contrastive Estimation}

\usepackage[output-decimal-marker=\text{.}, range-phrase = \text{--}]{siunitx}
\makeatletter
\newcommand{\@giventhatstar}[2]{$\left(#1\,\middle|\,#2\right)$}
\newcommand{\@giventhatnostar}[3][]{#1(#2\,#1|\,#3#1)}
\newcommand{\giventhat}{\@ifstar\@giventhatstar\@giventhatnostar}
\makeatother

\newcommand{\dist}[3]{#1\giventhat{#2}{#3}}
\newcommand{\expnumber}[2]{{#1}\mathrm{e}{#2}}
\DeclarePairedDelimiter\floor{\lfloor}{\rfloor}

\DeclarePairedDelimiterX{\infdivx}[2]{[}{]}{%
  #1\,\delimsize\|\,#2%
}
\newcommand{\kldiv}{\text{KL}\infdivx}

\newcommand{\code}[1]{\texttt{#1}}

\newenvironment{tightcenter}{%
  \setlength\topsep{0pt}
  \setlength\parskip{0pt}
  \begin{center}
}{%
  \end{center}
}

\begin{document}

\twocolumn[
\icmltitle{Sample and Predict Your Latent: Modality-free Sequential Disentanglement via Contrastive Estimation}

% It is OKAY to include author information, even for blind
% submissions: the style file will automatically remove it for you
% unless you've provided the [accepted] option to the icml2020
% package.

% List of affiliations: The first argument should be a (short)
% identifier you will use later to specify author affiliations
% Academic affiliations should list Department, University, City, Region, Country
% Industry affiliations should list Company, City, Region, Country

% You can specify symbols, otherwise they are numbered in order.
% Ideally, you should not use this facility. Affiliations will be numbered
% in order of appearance and this is the preferred way.
\icmlsetsymbol{equal}{*}

\begin{icmlauthorlist}
\icmlauthor{Ilan Naiman}{equal,yyy}
\icmlauthor{Nimrod Berman}{equal,yyy}
\icmlauthor{Omri Azencot}{yyy}
\end{icmlauthorlist}

\icmlaffiliation{yyy}{Department of Computer Science, Ben-Gurion University of the Negev, Beer-Sheva, Israel}
\icmlcorrespondingauthor{Ilan Naiman}{naimani@post.bgu.ac.il}
% \icmlcorrespondingauthor{Nimrod Berman}{bermann@post.bgu.ac.il}
% \icmlcorrespondingauthor{Omri Azencot}{azencot@cs.bgu.ac.il}

% You may provide any keywords that you
% find helpful for describing your paper; these are used to populate
% the "keywords" metadata in the PDF but will not be shown in the document
\icmlkeywords{TODO}

\vskip 0.15in
]

% \printAffiliationsAndNotice{}  % leave blank if no need to mention equal contribution
\printAffiliationsAndNotice{\icmlEqualContribution} % otherwise use the standard text.

\begin{abstract}

% (sequential) disentanglement; self-supervision; challenges; our observation; our approach; our results;
Unsupervised disentanglement is a long-standing challenge in representation learning. Recently, self-supervised techniques achieved impressive results in the sequential setting, where data is time-dependent. However, the latter methods employ modality-based data augmentations and random sampling or solve auxiliary tasks. In this work, we propose to avoid that by generating, sampling, and comparing empirical distributions from the underlying variational model. Unlike existing work, we introduce a self-supervised sequential disentanglement framework based on contrastive estimation with \emph{no} external signals, while using common batch sizes and samples from the latent space itself. In practice, we propose a unified, efficient, and easy-to-code sampling strategy for semantically similar and dissimilar views of the data. We evaluate our approach on video, audio, and time series benchmarks. Our method presents state-of-the-art results in comparison to existing techniques. The code is available at \href{https://github.com/azencot-group/SPYL}{GitHub}.

\end{abstract}

\vspace{-6mm}
\section{Introduction}

% no labels, unsupervised learning, self-supervised learning
One of the main challenges for modern learning frameworks in tackling new tasks is the lack of high-quality real-world labeled data. Unfortunately, labeling massive amounts of data is a time-consuming process that typically requires expert knowledge. Unsupervised learning is a modeling paradigm for learning without labels, and thus it gained increased attention in recent years~\cite{sohl2015deep}. Recent approaches utilize the inputs as supervisory signals~\cite{chen2020simple} and use pretext tasks~\cite{misra2020self}, yielding highly-competitive self-supervised learning (SSL) frameworks~\cite{caron2020unsupervised}. The goal of this paper is to study the effect of a novel SSL approach on sequential disentanglement problems.

% disentanglement, representation learning, sequential disentanglement, variational autoencoders, 
Data disentanglement is related to representation learning, where semantic latent representations are sought, to be used in various downstream tasks. A common goal in sequential disentanglement is the factorization of data to time-invariant (i.e., static) and time-variant (i.e., dynamic) features~\cite{hsu2017unsupervised}. Most sequential disentanglement approaches for arbitrary data modalities such as video, audio, and time series are unsupervised, modeling the task via variational autoencoders (VAE)~\cite{hsu2017unsupervised, yingzhen2018disentangled, han2021disentangled}. Effectively, the static and dynamic factors are obtained via separate posterior distributions.

% SSL + SD; shortcomings + examples: modality-dependent supervision,
Self-supervised learning appeared only recently in sequence disentanglement works via supervisory signals, pretext tasks, and contrastive estimation. However, existing SSL introduces several shortcomings as it depends on the underlying modality. In this work, modality refers to the properties of the data or task. For instance, \cite{zhu2020s3vae} design auxiliary tasks per data type, e.g., predict silence in audio segments or detect a face in an image. Similarly, \cite{bai2021contrastively} require positive and negative samples with respect to the input, i.e., same-class and different-class examples, respectively. In practice, positive views are obtained via data-dependent data augmentation transformations such as rotations and cropping, whereas negative views are selected randomly from the batch. To increase the variability in the batch, common solutions address these issues by increasing the batch or creating a memory bank, resulting in high memory costs. In this work, we refer to the above approaches as modality-based supervision methods, and we argue that they can be avoided if the underlying model is generative.

% our approach, our observations, VAE supports comparison of and sampling from distributions, evaluate on benchmark tasks
To alleviate the above disadvantages, we design a novel sampling technique, yielding a new contrastive learning framework for disentanglement tasks of \emph{arbitrary} sequential data that is based on the following insights. First, variational autoencoders naturally support the comparison of empirical distributions and their sampling. Second, we observe that a sample may be contrasted with its subsequent VAE prediction, leading to an increase in batch variability while keeping its size fixed. Based on these observations, we will show that we generate good positive and negative views. We evaluate our method on several challenging disentanglement problems and downstream tasks, and we achieve beyond state-of-the-art (SOTA) performance in comparison to several strong baseline methods.

\paragraph{Contributions.} Our main contributions are listed below.
\vspace{-1mm}
\begin{enumerate} 
    \itemsep-.1em
    \item We observe that VAEs allow to compare and construct empirical distributions while facilitating a contrastive evaluation of samples and their prediction.
    \item We propose a novel similarity and sampling technique that exploits inherent properties of VAEs, is modality-free, and it yields good views during training.
    \item We present SOTA results on video, audio, as well as time series data; our empirical evaluation includes data and task modalities for which existing SSL approaches are not necessarily effective, such as the Physionet ICU dataset~\cite{goldberger2000physiobank}.
\end{enumerate}

\section{Related Work}

% contrastive estimation: classic, modern, theoretical vs. empirical results
\paragraph{Contrastive learning.} Drawing two semantic views of the same object closer already appeared in signature verification systems~\cite{bromley1993signature}. However, issues such as \emph{catastrophic collapse} were identified in~\cite{chopra2005learning} i.e., learned views may ``collapse'' to a constant function, yielding zero similarity but alas, non-useful representations. To avoid collapse, they draw similar inputs closer, while separating dissimilar inputs via a contrastive term. Gutmann et al.~\yrcite{gutmann2010noise} suggested a popular and theoretically-sound framework to discriminate between observed data and artificial noise, known as noise contrastive estimation. In Contrastive Predictive Coding (CPC)~\cite{oord2018representation}, the authors generate robust representations by contrasting the predicted next frame. These results have been integrated in vision tasks with frameworks such as SimCLR~\cite{chen2020simple} and InfoMin~\cite{tian2020makes}. Recent works suggest regularizing the loss~\cite{tsai2021self} and utilizing contrastive learning losses~\cite{zhu2021improving}.

\vspace{-1mm}
\paragraph{Contrastive sampling.} Contrastive learning techniques require semantically similar samples, as well as semantically dissimilar samples. These examples are often referred to as positive and negative examples, respectively. Often, positive samples are obtained via data augmentation tools, whereas negative samples are selected randomly~\cite{le2020contrastive}. A recent study~\cite{tian2020makes} devises optimality conditions on positive views. Specifically, they show that one should reduce the mutual information (MI) between views while keeping task-relevant information intact. We now describe several sampling approaches for positive and negative views that can be considered as reducing MI.

As mentioned above, the majority of approaches use data augmentation for positive sampling~\cite{bachman2019learning, chen2020simple}. In~\cite{sermanet2018time}, positive examples are obtained from video frames of different views for the same action, e.g., pouring coffee into a cup. A related idea appeared in~\cite{han2019video}, where given a collection of videos, they generate samples by considering different videos, and different locations/times within a video. \cite{ho2020contrastive} use positive adversarial samples to further enhance the effect of contrastive learning.

While significant attention was given to positive sampling techniques, recent approaches focus on negative sampling that goes beyond random selection from the batch~\cite{doersch2017multi} or from a memory bank~\cite{wu2018unsupervised, misra2020self, he2020momentum}. The main issue with random sampling is that it may yield negative examples which are actually positive, an issue known as sampling bias~\cite{chuang2020debiased}. To address the latter problem, \citet{kalantidis2020hard, robinson2020contrastive} construct negative samples by measuring their similarity to the current sample. Recently, \citet{ge2021robust} generate negative examples with superfluous features, and similarly, \citet{huynh2022boosting} aim at discarding semantic information from negative samples. \citet{ash2021investigating} studied the effect of the number of negative samples on performance. Finally, a few techniques showed impressive results with no negative examples at all~\cite{chuang2020debiased, grill2020bootstrap}.

\vspace{-1mm}
\paragraph{Disentanglement methods.} 

Separating the underlying factors of variation is a well-established research problem on static image data~\cite{kulkarni2015deep, higgins2016beta, kim2018disentangling, chen2018isolating}. Disentanglement of sequential data is an emerging field, and it focuses on data factorization to static and dynamic factors. \citet{hsu2017unsupervised} introduced unsupervised disentanglement of sequential data via an LSTM-based model on audio data. Later, \citet{yingzhen2018disentangled} suggested DSVAE using a similar LSTM architecture while adding a heuristic in which the dynamic features' dimension is small compared to the static features' size. Further, \citet{tulyakov2018mocogan} proposed an adversarial setup. S3VAE~\cite{zhu2020s3vae} improves DSVAE by adding mutual information penalties on the relation between the static and dynamic features and the input, and in addition, they used auxiliary signals. \citet{han2021disentangled} also suggested improving DSVAE by replacing the Euclidean distance with a Wasserstein distance. \citet{tonekaboni2022decoupling} use a VAE model to disentangle arbitrary time series data. C-DSVAE~\cite{bai2021contrastively} includes a contrastive estimation of the mutual information losses introduced by S3VAE. They employ data augmentations for contrastive loss estimation, using a similar architecture as S3VAE and DSVAE. Recent work by~\citet{berman2023multifactor} developed structured Koopman autoencoders to promote multifactor disentanglement of the sequential data to more than two semantic components. Our work builds on the architecture and objective of C-DSVAE while overcoming some of its shortcomings. Specifically, we design a simple framework for sampling good positive and negative samples. Our approach is modality-free, i.e., it does not depend on the data domain (video, audio, or time series), nor does it depend on the task (e.g., images of faces or letter images).

% self-supervision and contrastive estimation in sequential disentanglement
\paragraph{Contrastive disentanglement.} Several works considered contrastive estimation in the context of disentanglement of latent factors~\cite{lin2020infogan, li2021disentangled, wang2021self}. Here, we focus on disentanglement of sequential data. For instance, \citet{wei2022unsupervised} employ a contrastive triplet loss for unsupervised video domain adaptation. Self-supervision in sequential disentanglement of arbitrary data appeared only recently. \citet{zhu2020s3vae} utilize auxiliary tasks and supervisory signals, whereas \citet{bai2021contrastively} use contrastive estimation, following the standard augmentation and random sampling for constructing positive and negative examples, respectively, and using the \code{infoNCE} loss.

\section{Background}
\label{sec:background}

\paragraph{Problem formulation.} Given a dataset $\mathcal{D} = \{ x_{1:T}^j \}_{j=1}^N$ of time series sequences $x_{1:T} = \{ x_1, \dots, x_T \}$ where the index $j$ is omitted for brevity, our goal is to find a posterior distribution $\dist{p}{s, d_{1:T}}{x_{1:T}}$ of \emph{disentangled} static and dynamic latent representations, $s$ and $d_{1:T}$ respectively, such that $x_{1:T} \sim \dist{p}{x_{1:T}}{s, d_{1:T}}$. We elaborate below on the constraints and assumptions related to the factors and data.

\paragraph{Probabilistic modeling.} Our discussion follows closely existing works such as~\cite{yingzhen2018disentangled, bai2021contrastively}. The static factor $s$ and dynamic factors $d_{1:T}$ are assumed to be independent, $x_i$ depends only on $s$ and $d_i$, and $d_i$ depends on the previous dynamic factors $d_{<i} = \{ d_1, \dots, d_{i-1}\}$. Under these assumptions, we consider the following joint distribution
\begin{equation} \label{eq:vae_prior}
    p(x_{1:T}, z) = \left[p(s) \prod_{i=1}^T \dist{p}{d_i}{d_{<i}} \right] \cdot \prod_{i=1}^T \dist{p}{x_i}{s, d_i} \ ,
\end{equation}
where $z = (s, d_{1:T})$. The prior distributions $p(s)$ and $\dist{p}{d_i}{d_{<i}}$ are taken to be Gaussian with $p(s) := \mathcal{N}(0, I)$ and $\dist{p}{d_i}{d_{<i}} := \mathcal{N}(\mu(d_{<i}), \sigma^2(d_{<i}))$. 

The posterior distribution $\dist{p}{s, d_{1:T}}{x_{1:T}}$ disentangles static from dynamic, and it is approximated via
\begin{equation} \label{eq:vae_posterior}
    \dist{q}{z}{x_{1:T}} = \dist{q}{s}{x_{1:T}} \prod_{i=1}^T \dist{q}{d_i}{d_{<i}, x_{\leq i}} \ ,
\end{equation}
i.e., $s$ is conditioned on the entire sequence, whereas $d_i$ depends on previous $d_j$ and inputs, and current inputs.

The variational autoencoder (VAE)~\cite{kingma2013auto} relates the prior and approximate posterior distributions in a regularized reconstruction loss. For mutually independent $s$ and $d_{1:T}$ this loss takes the following form,
\begin{align} \begin{split} \label{eq:vae_loss}
    \mathcal{L}_\text{VAE} &= \lambda_1 \, \mathbb{E}_{\dist{q}{z}{x_{1:T}}} \log \dist{p}{x_{1:T}}{z} \\
    &- \lambda_2 \, \kldiv{\dist{q}{s}{x_{1:T}}}{p(s)} \\
    &- \lambda_3 \, \kldiv{\dist{q}{d_{1:T}}{x_{1:T}}}{p(d_{1:T})} \ ,
\end{split} \end{align}
where $\kldiv{q}{p}$ is the Kullback--Leibler divergence that computes the distance between distributions $q$ and $p$, and $\lambda_1, \lambda_2, \lambda_3 \in \mathbb{R}^+$ are weight hyperparameters.

In practice, the likelihood $\dist{p}{x_i}{s,d_i}$ in Eq.~\eqref{eq:vae_prior}, $\dist{p}{d_i}{d_{<i}}$ and the terms $\dist{q}{s}{x_{1:T}}$ and $\dist{q}{d_i}{d_{<i}, x_{\leq i}}$ in Eq.~\eqref{eq:vae_posterior} are all obtained via separate LSTM modules. Sampling from the sequential distribution $\dist{p}{d_i}{d_{<i}}$ is achieved by using the mean and variance the LSTM outputs when feeding $d_{i-1}$, and similarly for $\dist{q}{d_i}{d_{<i}, x_{\leq i}}$. Finally, we use the mean squared error (MSE) for reconstruction in Eq.~\eqref{eq:vae_loss} and the KL terms are computed analytically. Further network architectural details are given in App.~\ref{app:model_arch}.

\paragraph{Mutual information disentanglement.} Similar to the catastrophic collapse observed in~\cite{chopra2005learning}, VAE models may produce non-informative latent factors~\cite{bowman2015generating}. In sequential disentanglement tasks, this issue manifests itself by condensing the static and dynamic information into $d_{1:T}$. An empirical heuristic has been partially successful in mitigating this issue, where a low-dimensional $d_i$ and a high-dimensional $s$ are used~\cite{yingzhen2018disentangled}, thus $d_i$ is less expressive by construction. However, a recent theoretical result~\cite{locatello2019challenging} shows that unsupervised disentanglement is impossible if no inductive biases are imposed on models and data. Thus, to alleviate these challenges, several existing works~\cite{zhu2020s3vae, han2021disentangled, bai2021contrastively} augmented model~\eqref{eq:vae_loss} with \emph{mutual information} terms.

The main idea in introducing mutual information (MI) terms is to separately maximize the relation in pairs $(s, x_{1:T})$ and $(d_{1:T}, x_{1:T})$, while minimizing the relation of $(s, d_{1:T})$. This idea is realized formally as follows~\cite{bai2021contrastively},
\begin{equation} \label{eq:mi_loss}
    \mathcal{L}_\text{MI} = \lambda_4 I_q(s; x_{1:T}) + \lambda_4 I_q(d_{1:T}; x_{1:T}) - \lambda_5 I_q(s; d_{1:T}) \ ,
\end{equation}
where $I_q(u; v) = \mathbb{E}_{q(u, v)} \log \frac{\dist{q}{u}{v}}{q(u)}$. Combining the above losses~\eqref{eq:vae_loss} and~\eqref{eq:mi_loss}, the disentanglement model reads
\begin{equation} \label{eq:dis_loss}
    \max_{p, q} \mathbb{E}_{x_{1:T}\sim p_\mathcal{D}} \mathcal{L}_\text{VAE} + \mathcal{L}_\text{MI} \ ,
\end{equation}
where $p_\mathcal{D}$ is the empirical distribution of the dataset $\mathcal{D}$. Bai et al.~\yrcite{bai2021contrastively} shows that problem~\eqref{eq:dis_loss} is a proper evidence lower bound (ELBO) of the log-likelihood of~\eqref{eq:vae_prior}.

Estimating the MI terms is not straightforward. A standard approach uses mini-batch weighted sampling (MWS)~\cite{chen2018isolating}. In contrast, \citet{bai2021contrastively} approximated MI terms via a contrastive estimation known as \code{infoNCE},
\begin{equation} \label{eq:info_nce}
     \mathcal{L}_\text{iNCE} = \log \frac{\phi(u, v^+)}{\phi(u, v^+) + \sum_{j=1}^M \phi(u, v^j)} \ ,
\end{equation}
where $u$ is either the static factor or the dynamic features, i.e., $u \in \{s, d_{1:T}\}$. The samples $v^+$ and $v^j$ correspond to positive and negative views with respect to $u$. For instance, if $u := s$, then the static features $v^+ := s^+$ are similar to $s$. The function $\phi(u, v)= \exp(u^T v / \tau |u| |v|)$ measures the similarity between examples $u$ and $v$, with $\tau=0.5$ being a temperature parameter~\cite{chen2020simple}. It has been shown that $I_q(u; x_{1:T}) \approx \mathcal{L}_\text{iNCE}(u)$ under relatively mild conditions~\cite{oord2018representation}. Our model architecture and objective function follow C-DSVAE~\cite{bai2021contrastively}, while significantly improving their contrastive estimation by proposing a novel sampling procedure as we detail below.

\begin{figure*}[ht]
    \centering
    \begin{overpic}[width=1\linewidth]{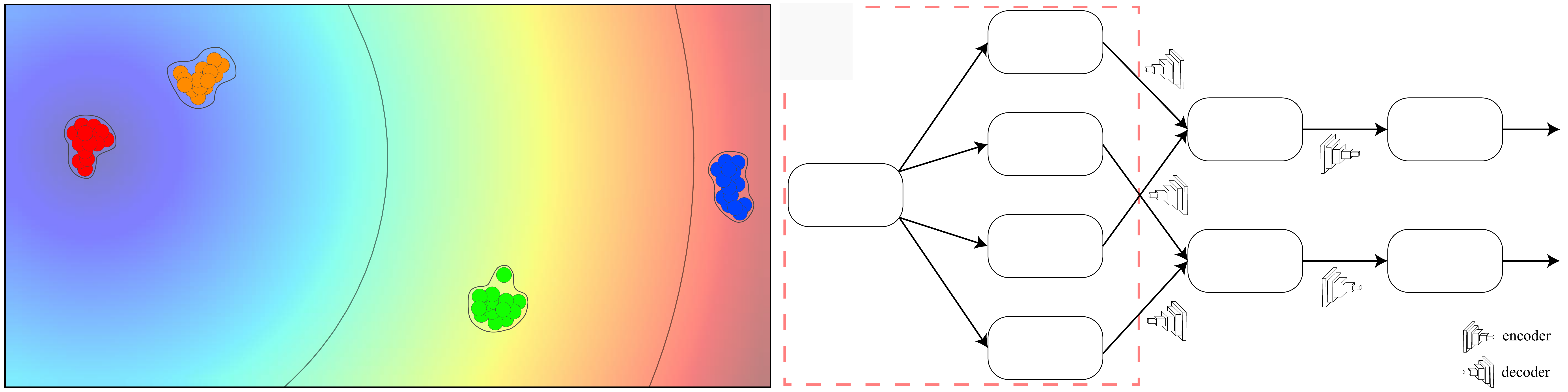}
        % left side
        \put(.5, 23){A} \put(3, 12){$s$} \put(15, 18){$S^+$} \put(46, 16){$S^-$}
        % right side (in boxes)
        \put(50, 23){B} \put(51.5, 12){$s, d_{1:T}$}
        \put(65.75, 21.5){$\tilde{s}^+$} \put(65, 15){$\tilde{s}^{-, j}$}
        \put(65, 8.5){$\tilde{d}_{1:T}$} \put(65, 2){$\tilde{d}_{1:T}^j$}
        \put(78, 16.25){$x_{1:T}^+$} \put(77.5, 8){$x_{1:T}^{-, j}$}
        \put(91, 16){$s^+$} \put(91, 7.5){$s^{-, j}$}
        % right side (on arrows)
        \put(59.5 ,20){$S^+$} \put(59.5 ,15.5){$S^-$}
        \put(59.5 ,9){$p$} \put(59.5 ,5){$p$}
        \put(73 ,16){$p$} \put(73 ,8){$p$}
        \put(85 ,18){$q$} \put(85 ,9.5){$q$}
        {\scriptsize \put(96 ,18){$\mathcal{L}_\text{iNCE}$} \put(96 ,9.5){$\mathcal{L}_\text{iNCE}$}}
    \end{overpic}
    \vspace*{-4mm}
    \caption{A) To generate positive and negative static views of $s$, we collect the closest $S^+$ and farthest $S^-$ distributions in the batch. We sample from these distributions using the reparameterization trick.  B) Unfortunately, samples from the batch have limited variation (dashed red rectangle), and thus we use our predictive sampling trick, generating samples by using the posterior of the sampled prior.}
    \label{fig:sampling_n_scheme}
\end{figure*}
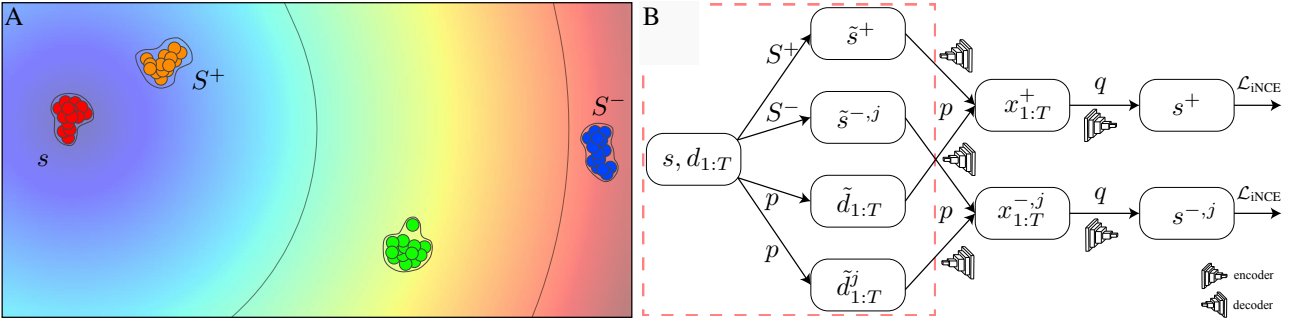

\section{Method}
\label{sec:method}

% existing sequential disentanglement and contrastive learning; sampling and contrastive estimation
Employing contrastive estimation~\eqref{eq:info_nce} within a sequential disentanglement framework requires positive and negative views of $s$ and $d_{1:T}$ for a given input $x_{1:T}$. In practice, the positive samples are obtained via \emph{modality-based} data augmentations such as cropping and color distortion for images, voice conversion for audio data, and shuffling of frames for general static augmentation. Negative views are obtained by randomly sampling from the batch. See, for instance, \cite{zhu2020s3vae, bai2021contrastively}. In this work, we argue that while random sampling and data augmentations with the \code{infoNCE} loss are popular tools for unsupervised learning \cite{chen2020simple, tian2020makes}, one should revisit the core components of sequential contrastive learning. We will show that existing practices for sampling of views and for increasing the batch variation can be improved.

% challenges of DA; data modality vs. task modality; (challenging modalities with no/scarce DA)
\paragraph{Shortcomings of views' sampling.} Creating semantically similar and dissimilar examples is a challenging problem. We distinguish between \emph{domain-modality} and \emph{task-modality} sampling. In general, we will use the following definition of $X$-modality. A method $Y$ is $X$-modality-dependent if $Y$ depends on the characteristics of $X$. For instance, image rotation ($Y$) is domain-modality-dependent ($X$) as it will probably be less effective for audio sequences. Similarly, cropping of images ($Y$) may not be effective for images of letters, and thus it is task-modality-dependent ($X$). Namely, task-modality-dependent approaches may require separate sampling methods for the same data domain. In summary, modality may mean multiple concepts depending on the particular context, including the format of the data or its statistical features. In general, we argue below that existing disentanglement approaches are modality-based.

Existing studies show that the particular choice of DA can significantly affect results \cite{chen2020improved, tian2020makes, zhang2022rethinking}. Even shuffling of frames which may seem robust, can yield wrong views in critical healthcare applications involving data with the vital measurements of a patient. In conclusion, DA may heavily depend on domain knowledge and task expertise. DA which falls into one of the categories above is referred to as \emph{modality-based} augmentations. To the best of our knowledge, the majority of data augmentation tools are modality-based. 

% challenges of randomness; sampling bias, large batch, memory bank, memory footprint, 
Constructing negative views may seem conceptually simpler in comparison to positive views, however, it bears its own challenges. Common methods select randomly from the dataset~\cite{doersch2017multi}. To reduce the sampling bias of false negative views~\cite{chuang2020debiased}, existing works suggest increasing batch sizes~\cite{chen2020simple} or using a memory bank~\cite{wu2018unsupervised}. Yet, the memory footprint of these methods is limiting. To conclude, both data augmentation and randomness should be avoided in the construction of positive and negative views.

\vspace{-1mm}

% VAE inherent support, empirical distributions; 
\paragraph{VAE-based sampling.} Motivated by the above discussion, we opt for an efficient modality-free sampling approach. We make the following key observation:
\begin{tightcenter}
\emph{variational autoencoders inherently support the formation, comparison, and sampling of empirical distributions}
\end{tightcenter}
Essentially, given a dataset $\mathcal{D} = \{x_{1:T}^j\}_{j=1}^N$ of time series sequences and model~\eqref{eq:vae_loss}, we can generate the individual posterior distributions $\{ \dist{q}{z^j}{x_{1:T}^j} \}_{j=1}^N$, compare them via the Kullback--Leibler divergence, and sample $z^j \sim \dist{q}{z^j}{x_{1:T}^j}$.

We denote by $x_{1:T}$ the input for which we seek a positive $x_{1:T}^+$ and several negative $x_{1:T}^{-,j}, j=1, \dots, M$, examples. Our discussion focuses on sampling static views $\{s^+, s^{-,j}\}$, however, a similar process can be performed for sampling dynamic features. Intuitively, $s^+$ is the factor such that the distance $\kldiv{\dist{q}{s}{x_{1:T}}}{\dist{q}{s^+}{x_{1:T}^+}}$ is minimal, where $x_{1:T}^+ \sim p_\mathcal{D}$. Similarly, $s^{-,j}$ are the features with maximal KL value. However, a subtle yet important aspect of views is the distinction between \emph{soft and hard} samples. Soft negative examples are those which contribute less to learning as they are too dissimilar to the current sample, whereas hard views are the semantically-dissimilar examples that are close in latent space to $x_{1:T}$ \cite{kalantidis2020hard, robinson2020contrastive}. How would one obtain good views with large batch variability given the above observation?

% use batch, notations: partially-trained posterior and KLD dist. mat., sort and sample: 1/3 pos., 2/3 neg.
To increase variation while avoiding memory banks and large batch sizes, we suggest to use the (non-increased) batch itself. Let $\dist{\tilde{q}_k}{s}{x_{1:T}}$ denote the partially-trained posterior after $k$ epochs of training. We denote by $D \in \mathbb{R}^{n \times n}$ the pairwise KL divergence distances matrix for a batch of size $n$, $\{ x_{1:T}^j \}_{j=1}^n$. Namely,
\begin{equation} \label{eq:kl_dist}
    D_{ij} := \kldiv{\dist{\tilde{q}}{s^i}{x_{1:T}^i}}{\dist{\tilde{q}}{s^j}{x_{1:T}^j}} \ ,
\end{equation}
where $i,j \in \{1,\dots,n\}$ and we omit the training epoch for brevity. For a particular example in the batch $x_{1:T}^i$, we generate good views based on the following heuristic. We sort the row $D_{i:}$ in ascending order, and we sample positive views from the \emph{first} third of distributions, whereas negative views are sampled from the \emph{last} third. We denote by $S^+(i) = \{ \dist{\tilde{q}}{s^j}{x_{1:T}^j} \}$ the set of positive distributions, and similarly, $S^-(i)$ holds the negative distributions. See Fig.~\ref{fig:sampling_n_scheme} for an illustration of these definitions. 

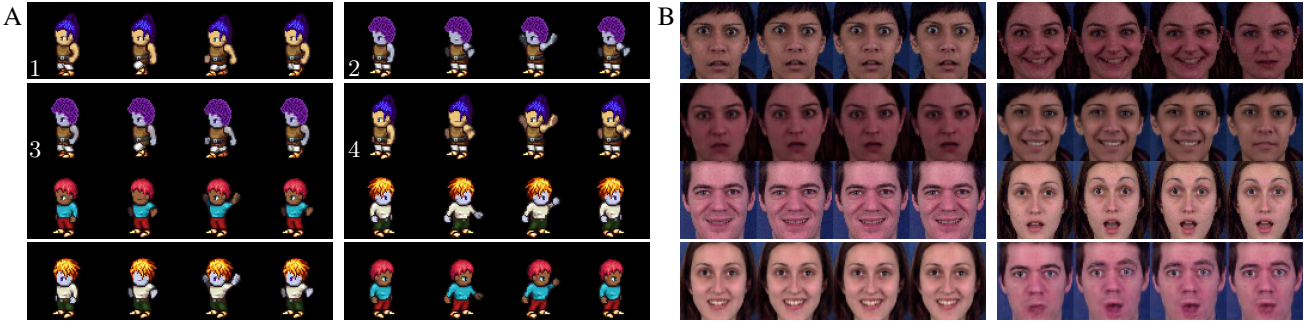
\begin{figure*}[t]
    \centering
    \begin{overpic}[width=1\linewidth]{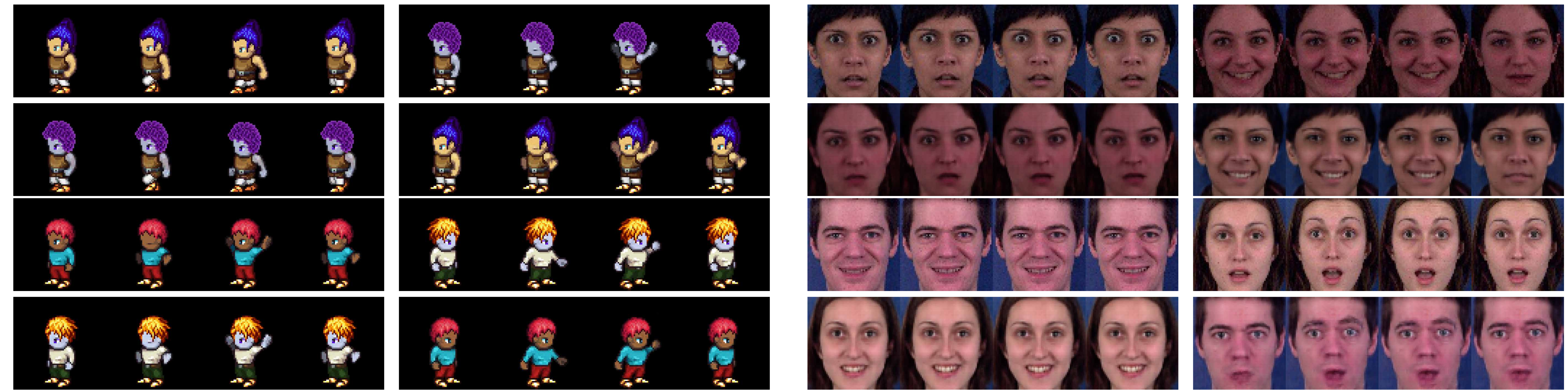}
        \put(-1, 23){A} \put(49.75, 23){B}
        \put(1, 19){{\footnotesize\color{white}$1$}} \put(25.75, 19){{\footnotesize\color{white}$2$}}
        \put(1, 12.75){{\footnotesize\color{white}$3$}} \put(25.75, 12.75){{\footnotesize\color{white}$4$}}
    \end{overpic}
    \vspace*{-4mm}
    \caption{Content and pose swap results in Sprites (A) and MUG (B) datasets. See the text for additional details.}
    \label{fig:swap_sprites_mug}
\end{figure*}

% insufficient deviation in batch;  existence of soft pos./neg.; how get soft to semi-soft pos. and semi-hard to hard neg. increase variability via predictive sampling trick, 
\paragraph{Predictive sampling trick.} Unfortunately, as usual batch sizes are relatively small, it may occur that variability is limited in the original batch. Notice that soft positive views always exist via the posterior of the sample itself, and soft negatives probably exist as well for moderate batch sizes, e.g., for $n=16, 32$. However, it is not clear whether hard views exist in the batch, and thus its variability may need to be increased. To improve variability, we introduce our \emph{predictive sampling trick}. 

% describe the process; dynamic: fix static, sample dynamic from prior; decode and encode; positive vs. negative sampling
Again, w.l.o.g. we focus on the setting of sampling static views of a given example $x_{1:T}$ with its static and dynamic features $s$ and $d_{1:T}$. To increase the variability in the views, our predictive sampling trick generates these examples from \emph{the posterior of the sampled prior}. For instance, to produce a positive static view, we denote $\tilde{s}^+ \sim S^+$. The dynamic features can be arbitrary, and thus we sample from the prior $\tilde{d}_{1:T} \sim p(d_{1:T})$. The positive instance $x_{1:T}^+$ is defined via $x_{1:T}^+ \sim \dist{p}{x_{1:T}^+}{\tilde{s}^+, \tilde{d}_{1:T}}$. We obtain the positive static view by sampling the posterior, i.e.,
\begin{equation}
    s^+ \sim \dist{q}{s^+}{x_{1:T}^+} \ .
\end{equation}
A similar process is used to compute $s^{-, j}, j=1,\dots,M$. These views $\{ s^+, s^{-, j} \}$ are utilized in $\mathcal{L}_\text{iNCE}(s, s^+, s^{-, j})$, see the diagram of our predictive sampling in Fig.~\ref{fig:sampling_n_scheme}B. We find that our views' heuristic and predictive sampling trick yield soft to semi-soft positive examples and semi-hard to hard negative examples, see Sec.~\ref{subsec:views_analysis}. For additional implementation details, see App.~\ref{app:impel_details}.

Our approach is based on the implicit assumption that the underlying model~\eqref{eq:vae_loss} encourages similar examples to be close and dissimilar views to be farther apart. Indeed, previous work on this model~\cite{yingzhen2018disentangled} showed this tendency when using large $s$ and small $d_i$. Thus, our approach can be viewed as promoting the natural tendency of the model to separate positive and negative views.

\section{Results}
\label{sec:results}

\subsection{Datasets and Methods}

In our evaluation, we consider several datasets of different modalities, and we compare our results with several state-of-the-art approaches. Specifically, we test on video datasets such as Sprites~\cite{reed2015deep} and MUG~\cite{5617662} containing animated cartoon characters and subjects performing facial expressions, respectively. Moreover, we also use the Jester dataset~\cite{materzynska2019jester} with videos of hand gestures, and the Letters corpus~\cite{letters} with handwritten text. For audio, we experiment with TIMIT~\cite{timit}. Finally, we also explore time series datasets including Physionet~\cite{goldberger2000physiobank} with individual medical records and Air Quality \cite{zhang2017cautionary} with measurements of multiple air pollutants. We compare our results to sequential disentanglement frameworks including MoCoGan~\cite{tulyakov2018mocogan}, FHVAE~\cite{hsu2017unsupervised}, DSVAE~\cite{yingzhen2018disentangled}, R-WAE~\cite{han2021disentangled}, S3VAE~\cite{zhu2020s3vae}, SKD~\cite{berman2023multifactor}, C-DSVAE~\cite{bai2021contrastively}, and GLR~\cite{tonekaboni2022decoupling}. See App.~\ref{app:exp_setup} for details.

\subsection{Hyperparameters}

To control the contribution of each loss component we add a $\lambda_1$ coefficient to the reconstruction loss, a $\lambda_2$ to the static KL term, a $\lambda_3$ to the dynamic KL term, and finally $\lambda_4, \lambda_5$ to the contrastive terms. The hyperparameter $\lambda_1$ is tuned over $\{1, 2.5, 5, 10\}$, $\lambda_2$ is tuned over $\{1, 3, 5, 7, 9\}$, and $\lambda_4$ and $\lambda_5$ are tuned over $\{0.1, 0.5, 1, 2.5, 5\}$ while $\lambda_3$ is fixed to $1$. We used Adam optimizer~\cite{kingma2014adam} with the learning rate chosen from $\{0.001, 0.0015, 0.002\}$. The static and dynamic features' dimensions are selected from $\{128, 256\}$ and $\{32, 64\}$, respectively. These dimensions are similar or sometimes smaller in comparison to all other benchmark models such as C-DSVAE, S3VAE, DSVAE, R-WAE. We highlight that tuning multiple hyperparameters is often challenging. Hence, we utilize automatic tuning tools, using $5$ to $10$ runs for each dataset. The hyperparameters for each task and dataset are given in Tab.~\ref{tab:hypers} in the Appendix. All the tasks were trained for at most $600$ epochs.

\subsection{Qualitative Results}

We begin our evaluation with qualitative examples showing the disentanglement capabilities of our approach in Fig.~\ref{fig:swap_sprites_mug}. Specifically, given source and target samples, $x_{1:T}^\text{src}, x_{1:T}^\text{tgt}$, we swap the static and dynamic features between the source and the target. In practice, swapping the content (static) information corresponds to generating an image with factors $(s^\text{tgt}, d_{1:T}^\text{src})$, i.e., fix the source dynamics and use the static factor of the target. For instance, a perfect content swap in Sprites yields different characters with the same pose. The opposite swap of pose (dynamic) information is obtained with $(s^\text{src}, d_{1:T}^\text{tgt})$. Fig.~\ref{fig:swap_sprites_mug} shows two separate examples of Sprites and MUG (rows $1,2$ and rows $3,4$), where each example is organized in blocks of four panels. For instance on the Sprites example, panel no. $1$ is the source and panel no. $2$ is the target. Panel no. $3$ shows a content (static) swap, and Panel no. $4$ shows a pose (dynamic) swap.

\subsection{Quantitative Results: Common Benchmarks}
\label{subsection:common_bench}

\setlength{\tabcolsep}{2pt}
\begin{table*}[!t]
    \caption{Disentanglement metrics on Sprites, MUG, and TIMIT. Results with standard deviation appear in \ref{app:table_1_expand}.}
    \label{tab:gen_sprites_mug_timit}
    % \fontsize{7.5}{9}
    
    \centering
    \footnotesize
        \begin{tabular}[t]{lcccc|cccc}
            \toprule
            & \multicolumn{4}{c|}{Sprites} & \multicolumn{4}{c}{MUG} \\
            Method & Acc$\uparrow$ & IS$\uparrow$ & $H(y|x){\downarrow}$ & $H(y){\uparrow}$ & Acc$\uparrow$ & IS$\uparrow$ & $H(y|x){\downarrow}$ & $H(y){\uparrow}$ \\
            \midrule
            MoCoGAN & $92.89\%$ & $8.461$ & $0.090$ & $2.192$ & $63.12\%$ & $4.332$ & $0.183$ & $1.721$ \\
            DSVAE & $90.73\%$ & $8.384$ & $0.072$ & $2.192$ & $54.29\%$ & $3.608$ & $0.374$ & $1.657$ \\
            R-WAE & $98.98\%$ & $8.516$ & $0.055$ & $\boldsymbol{2.197}$ & $71.25\%$ & $5.149$ & $0.131$ & $1.771$ \\
            S3VAE & $99.49\%$ & $8.637$ & $0.041$ & $\boldsymbol{2.197}$ & $70.51\%$ & $5.136$ & $0.135$ & $1.760$ \\
            SKD & $\boldsymbol{100\%}$ & $\boldsymbol{8.999}$ & $\boldsymbol{\expnumber{1.6}{-7}}$ & $\boldsymbol{2.197}$ & $77.45\%$ & $\boldsymbol{5.569}$ & $\boldsymbol{0.052}$ & $1.769$ \\
            C-DSVAE & $99.99\%$ & $8.871$ & $0.014$ & $\boldsymbol{2.197}$ & $81.16\%$ & $5.341$ & $0.092$ & $1.775$ \\
            \midrule
            Ours & $\boldsymbol{100\%}$ & $8.942$ & $0.006$ & $\boldsymbol{2.197}$ & $\boldsymbol{85.71\%}$ & $5.548$ & $0.066$ & $\boldsymbol{1.779}$ \\
            \bottomrule
        \end{tabular}
        % \hfill
        \hfill
        \begin{tabular}[t]{lcc}
            \toprule
            & \multicolumn{2}{c}{TIMIT} \\
            Method & static EER$\downarrow$ & dynamic EER $\uparrow$ \\
            \midrule
            FHVAE & $5.06\%$ & $22.77\%$  \\
            DSVAE & $5.64\%$ & $19.20\%$  \\
            R-WAE & $4.73\%$ & $23.41\%$ \\
            S3VAE & $5.02\%$ & $25.51\%$  \\
            SKD   & $4.46\%$ & $26.78\%$  \\
            C-DSVAE & $4.03\%$ & $31.81\%$  \\
            \midrule
            Ours & $\boldsymbol{3.41\%}$ & $\boldsymbol{33.22 \%}$ \\
            \bottomrule
        \end{tabular}
\end{table*}

\paragraph{Image data.} Similar to previous work~\cite{zhu2020s3vae, bai2021contrastively}, we test our model disentanglement and generative abilities on the Sprites and MUG datasets, and we compare our results with state-of-the-art (SOTA) methods. The evaluation protocol takes a sample $x_{1:T}$ with its static and dynamic factors, $s$ and $d_{1:T}$, and generates a new sample $\tilde{x}_{1:T}$ with the original dynamic features, and a new static component sampled from the prior, $\tilde{s} \sim p(s)$. Ideally, we expect that $x_{1:T}, \tilde{x}_{1:T}$ share the dynamic labels, e.g., happy in MUG, whereas, their static classes match with probability close to random guess. To verify that this is indeed the case, we use a pre-trained classifier to predict the dynamic label of $\tilde{x}_{1:T}$, and we compare it to the true label of $x_{1:T}$. 

We utilize these labels on several different error metrics: label accuracy (Acc), inception score (IS) that estimates the generator performance, intra-entropy $H(y|x)$ that shows how confident the classifier is regarding its prediction, and inter-entropy $H(y)$ that measures diversity in generated samples, see also App.~\ref{app:exp_setup}. Our results are provided in Tab.~\ref{tab:gen_sprites_mug_timit}, alongside the results of previous SOTA approaches. The arrows $\uparrow, \downarrow$ next to the metrics denote which metric is expected to be higher or lower in value, respectively. Notably, our method outperforms existing work on Sprites and MUG datasets with respect to all metrics. Finally, one may also consider the opposite test where the static factor is fixed and the dynamic features are sampled. However, the SOTA methods achieve near perfect accuracy in this setting, and thus, we do not show these results here.

\paragraph{Audio data.} Another common benchmark demonstrates the effectiveness of sequential disentanglement frameworks on a different data modality~\cite{hsu2017unsupervised, yingzhen2018disentangled}. Specifically, we consider speaker verification on the TIMIT dataset. The main objective is to distinguish between different speakers, independently of the text they read. For a sample $x_{1:T}$, we expect that its static factor $s$ represents the speaker identity, whereas $d_{1:T}$ should not be related to that information. We use the Equal Error Rate (EER) metric where we compute the cosine similarity between all $s$ instances and independently for $d_{1:T}$ instances. Two static vectors encode the same speaker if their cosine similarity is higher than a threshold $\epsilon \in [0, 1]$, and different speakers otherwise. The threshold $\epsilon$ needs to be calibrated to receive the EER~\citep{chenafa2008biometric}. Tab.~\ref{tab:gen_sprites_mug_timit} shows that our approach improves SOTA results by a margin of $0.62\%$ and $1.41\%$ for the static and dynamic EER. For additional info on this benchmark, see App.~\ref{app:timit}.

\paragraph{Time series data.} Recently, \cite{tonekaboni2022decoupling} explored their approach on downstream tasks with time series data. Sequential information different from image and audio is an ideal test case for our framework as we lift the dependency on data augmentation (DA) techniques. Indeed, while DA is common for image/audio data, it is less available for arbitrary time series data. We follow the evaluation setup in~\cite{tonekaboni2022decoupling} to study the latent representations learned by our method. Specifically, we used an encoder and a decoder to compute codes of consecutive time series windows, and we extract these codes on non-stationary datasets such as Physionet and air quality.

We consider the following tasks: 1. prediction of the risk of in-hospital mortality, and 2. estimation of the average daily rain level. For each task, we train a simple RNN classifier in which we utilize the latent representations from the above autoencoder. For comparison, Tonekaboni et al.~\yrcite{tonekaboni2022decoupling} used C-DSVAE without data augmentation and thus with no contrastive estimation losses. However, as noted in the above paragraph, our approach does not have this limitation, and thus we can utilize the entire model~\eqref{eq:dis_loss}. Tab.~\ref{tab:ts_res} shows the results on the mortality rate and daily rain downstream tasks. Our method performs on par with GLR on the mortality rate task, and it comes second for daily rain estimation. However, it is important to emphasize that GLR was designed specifically for time series data with statistical properties as in Physionet and Air Quality datasets. In contrast, our method is not tuned to specifically handle time series data, and it can work on multiple data modalities such as video, audio, and time series data. Further, In our experimental setup, we were not able to re-produce the baseline results for the daily rain task. We leave this direction for further exploration. For more details regarding the evaluation setup and tasks, we refer to~\cite{tonekaboni2022decoupling}.

\subsection{Quantitative Results: New Benchmarks}
\label{exp:latent_features_exp}

The standard sequential disentanglement benchmark tests include classification of conditionally generated images and speaker verification. Here, we propose a new benchmark that quantifies the quality of the learned representations. For this evaluation, we consider the MUG dataset, and two challenging video datasets with hand writing (Letters) and hand gestures (Jesters). In this experiment, we explore a common framework to evaluate the disentangled codes. First, we compute the static $\{ s^j \}$ and dynamic $\{ d_{1:T}^j \}$ codes of the test set. Then, we define train and test sets via an $80-20$ split of the test set, and we train four classifiers. The first classifier takes $s$ vectors as inputs, and it tries to predict the static label. The second classifier takes $s$ and it predicts the dynamic label. Similarly, the third classifier takes $d_{1:T}$ and predicts the static label, and the fourth classifier takes $d_{1:T}$ and predicts the dynamic label. An ideal result with input $s$ is a prefect classification score in the first classifier, and a random guess in the second classifier. Additional details on this experiment appear in~\ref{app:ltnt_features_task}. Our results are summarized in Tab.~\ref{tab:dis_met_cls_task}, where we outperform C-DSVAE often by a large gap in accuracy. The Jesters dataset does not include static labels, and thus we only have partial results. Further, this dataset is extremely challenging due to low-quality images and complex gestures, and currently, C-DSVAE and our approach obtain low scores, where our approach attains $>7\%$ improvement over a random guess. These results can be improved by integrating recent VAEs \cite{razavi2019generating, vahdat2020nvae}, as we observe low quality reconstruction, which may effect disentanglement abilities.

\begin{table}[b]
    \vspace{-3mm}
    \caption{Error metrics on Physionet and Air quality datasets}
    \label{tab:ts_res}
    \centering
    \footnotesize
    \begin{tabular}{lcccc}
        \toprule
        & \multicolumn{2}{c}{ICU Mortality Prediction} & Avg. Daily Rain \\
        Method & AUPRC & AUROC & MAE \\
        \midrule
        VAE        & $0.157 \pm 0.053$ & $0.564 \pm 0.044$ & $1.831 \pm 0.005$ \\
        GPVAE      & $0.282 \pm 0.086$ & $0.699 \pm 0.018$ & $1.826 \pm 0.001$ \\
        C-DSVAE     & $0.158 \pm 0.005$ & $0.565 \pm 0.007$ & $\boldsymbol{1.806 \pm 0.012}$ \\
        GLR        & $0.365 \pm 0.092$ & $0.752 \pm 0.011$ & $1.824 \pm 0.001$ \\
        \midrule
        Ours       & $\boldsymbol{0.367 \pm 0.015}$ & $\boldsymbol{0.764 \pm  0.040}$ & $1.823 \pm 0.001$  \\
        \bottomrule
    \end{tabular}
\end{table}

\begin{table*}
    \caption{Downstream classification task on latent static and dynamic features. Results with standard deviation appear in \ref{app:ltnt_features_task}.}
    \label{tab:dis_met_cls_task}
    
    \centering
    \footnotesize
    \begin{tabular}[t]{ll|ccc|ccc}
        \toprule
        & & \multicolumn{3}{c|}{Static features} & \multicolumn{3}{c}{Dynamic features} \\
        Dataset & Method  & Static L-Acc $\uparrow$ & Dynamic L-Acc $\downarrow$ & Gap $\uparrow$ & Static L-Acc  $\downarrow$ & Dynamic L-Acc $\uparrow$ & Gap $\uparrow$ \\
        \midrule
        \multirow{3}{*}{MUG} & random & $1.92 \%$ & $16.66\%$ & - & $1.92 \%$ & $16.66\%$ & - \\
        & C-DSVAE    & $\boldsymbol{98.75\%} $ & $76.25\% $ & $22.25\%$ & $26.25\%$ & $82.50\%$ & $56.25\%$ \\
        & Ours       & $98.12\% $ & $\boldsymbol{68.75\%}$ & $\boldsymbol{29.37\%}$ & $\boldsymbol{10.00\%}$ & $\boldsymbol{85.62\%}$ & $\boldsymbol{75.62\%}$ \\
        \midrule
        \multirow{3}{*}{Letters} & random  & $1.65 \%$ & $3.84 \%$ & - & $1.65 \%$ & $3.84 \%$ & - \\
        & C-DSVAE    & $95.47\%$ & $13.0\%$ & $82.47\%$ & $\boldsymbol{2.79\%}$ & $66.35\% $ & $63.56\%$ \\
        & Ours       & $\boldsymbol{100\%} $ & $\boldsymbol{12.16\%}$ & $\boldsymbol{87.84\%}$ & $3.06\%$ & $\boldsymbol{69.75\%}$ & $\boldsymbol{66.69\%}$ \\
        \midrule
        \multirow{3}{*}{Jesters}  & random  & - & - & - & - & $20\%$ & - \\
        & C-DSVAE    & - & - & - & - & $21.88\%$ & - \\
        & Ours       & - & - & - & - & $\boldsymbol{27.70}\%$ & - \\
        \bottomrule
    \end{tabular}
\end{table*}

% \vspace{-1mm}
\subsection{Analysis of Positive and Negative Views}
\label{subsec:views_analysis}

\begin{figure*}[t]
    \centering
    \begin{overpic}[width=0.95\linewidth]{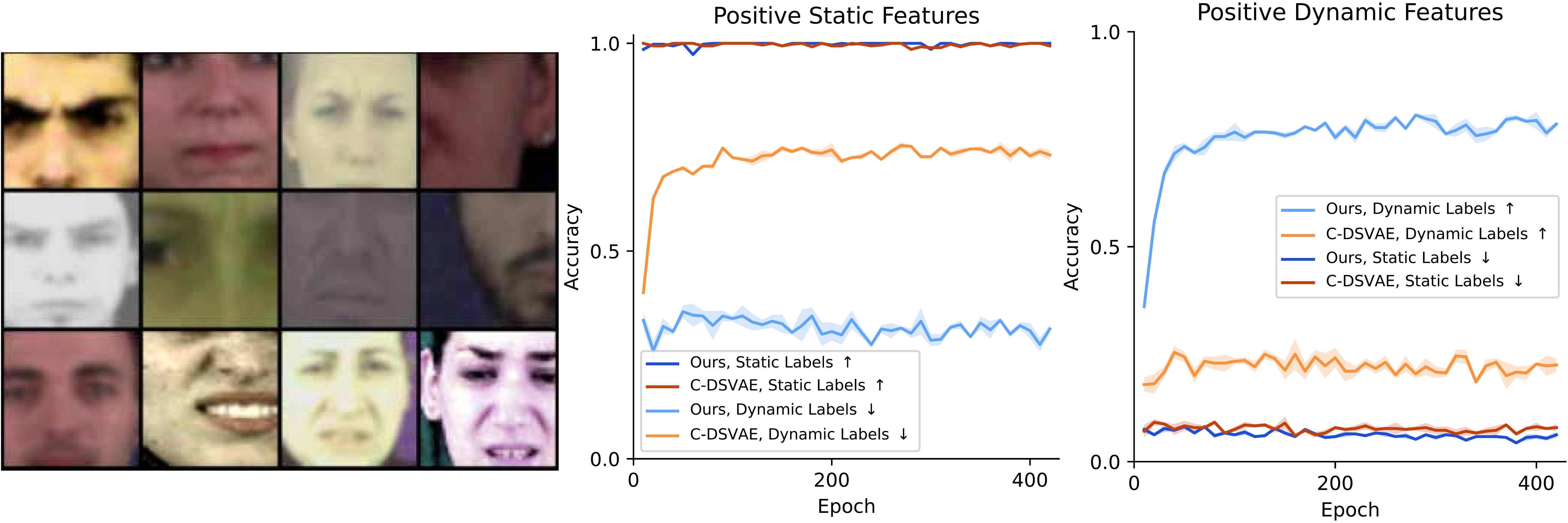}
        \put(.5, 32){A} \put(36, 32){$B$} \put(68, 32){$C$}
    \end{overpic}
    \vspace*{-3mm}
    \caption{We present randomly selected views of SimCLR on MUG (A). In addition, we compare the quality of views obtained with C-DSVAE and our approach when classifying the static and dynamic labels (B, C). See the text for more details.}
    \label{fig:views_analysis}
\end{figure*}

Sec.~\ref{sec:background} details how to incorporate contrastive learning in a sequential disentanglement setting, and in Sec.~\ref{sec:method}, we list some of the challenges such as sampling wrong positive and negative views. Here, we would like to empirically compare the views generated by C-DSVAE and our approach. For instance, we show a qualitative example in Fig.~\ref{fig:views_analysis}A of views used in C-DSVAE and obtained with SimCLR~\cite{chen2020simple}, where positive dynamic examples are generated via e.g., color distortion while supposedly keeping the dynamic features fixed. Unfurtunately, not all DA preserve the facial expressions. Beyond these qualitative examples, we also adapt the analysis \cite{tian2020makes} as detailed below.

Let $x_{1:T} \sim p_\mathcal{D}$ denote a data sample with its static and dynamic factors, $(s, d_{1:T})$. A positive static example $x_{1:T}^+$ is generated using the pair $(s^+, \tilde{d}_{1:T})$ where $s^+$ is similar to $s$, and $\tilde{d}_{1:T}$ is different from $d_{1:T}$. In the opposite case of a positive dynamic sample, the dynamic features $d_{1:T}^+$ are similar to $d_{1:T}$ and $\tilde{s}$ is different from $s$. Following Tian et al.~\yrcite{tian2020makes}, a good view is such that the mutual information $I_q(s^+; y)$ is high, whereas $I_q(\tilde{d}_{1:T} ; y)$ is low, where $y$ is the task label. For example, the identity of the person is kept, while its facial expression has changed. To estimate these MI terms, we use the latent codes in classification tasks as in Sec.~\ref{exp:latent_features_exp} where the static and dynamic factors are predicted. Namely, we use $s^+$ to predict the static labels, and similarly, we use $\tilde{d}_{1:T}$ to predict the dynamic labels. Good views will yield high static classification scores and low dynamic classification scores. 

We show in Fig.~\ref{fig:views_analysis}B the classification results when using $(s^+, \tilde{d}_{1:T})$ with C-DSVAE and with our method, and Fig.~\ref{fig:views_analysis}C shows the opposite case, i.e., $(\tilde{s}, d_{1:T}^+)$. For both plots, blue curves are related to our approach and red curves correspond to C-DSVAE. We focus on the test which uses $(s^+, \tilde{d}_{1:T})$, Fig.~\ref{fig:views_analysis}B. The blue and red curves show the classification accuracy when using $s^+$ to predict the static label, and thus, they should be high. In contrast, the light blue and orange curves arise from using $\tilde{d}_{1:T}$ to predict the dynamic labels, and thus, they should be close to a random guess for semi-hard views ($16.66\%$ in MUG). However, the orange curve is around $70\%$, whereas the light blue attains $\approx 30\%$. These results indicate that our views are semi-hard as they yield accuracy results close to a random guess. In the opposite scenario, Fig.~\ref{fig:views_analysis}C, we use the pair $(\tilde{s}, d_{1:T}^+)$ with different static and similar dynamic factors. Here, the blue and red curves should be close to a random guess ($1.92\%$), and the light blue and orange plots should present high accuracy values. However, the orange curve presents $\approx 25\%$ accuracy, whereas ours is around $70\%$. We conclude that our dynamic features better preserve the underlying action. Additional analysis and results are provided in App.~\ref{app:latent_emb}.

\subsection{Ablation Study: Negative Views Sampling}
\label{subsec:ablation}

The previous evaluation in Sec.~\ref{subsec:views_analysis} focused on the quality of positive views. Here, we explore the effect of utilizing various negative sampling rules. Ultimately, we would like to empirically motivate the heuristic we introduce in Sec.~\ref{sec:method} where we propose to only consider $33.3\%$ of the farthest distributions as measured by the KL divergence distance. An inferior heuristic is one that produces confusing negative views, i.e., examples that are semantically similar to the current data, instead of being dissimilar. However, choosing ``right'' negative views is important to the overall behavior of the approach, and thus we explore other sampling policies in the following ablation study. For a meaningful comparison, we fix the hyperparameters of the approach, and we train several models that only differ in their selection strategy of the negative views. Let $n$ the number of inputs in the batch; we define a pool of size $\floor{n/3}$ of negative distributions taken from: 1) the middle third, 2) the farthest third, and 3) both middle and farthest thirds, and 4) random sampling. From these distributions we sample $2n$ views. We present in Tab.~\ref{tab:ablt_negatives} the results of our ablation study on the MUG and TIMIT datasets. For MUG, we show the accuracy score, and we display the EER gap in TIMIT. Notably, all sampling strategies attain SOTA results on these tasks, cf. Tab.~\ref{tab:gen_sprites_mug_timit}. However, the farthest third yields the best results consistently across tasks, and thus, these results support our heuristic. In App. \ref{experiments:thirds_exp}, we conduct an analysis that motivates and justify our heuristic by showing the similarity distribution of the thirds.

\begin{table}[b]
    \caption{Negatives Ablation Study.}
    \label{tab:ablt_negatives}
    
    \centering
    \footnotesize
    \begin{tabular}{lccc}
        \toprule
        Negatives Mode      & Acc MUG$\uparrow$     & EER gap TIMIT$\uparrow$ \\
        \midrule
        Random              & $84.18\%$             & $28.53\%$ \\
        Middle Third        & $84.43\%$             & $29.11\%$ \\
        Middle+Farthest     & $84.96\%$             & $29.53\%$ \\
        Farthest Third      & $\boldsymbol{85.71\%}$ & $\boldsymbol{29.81\%}$ \\
        \bottomrule
        \hline
    \end{tabular}
\end{table}

\section{Limitations}

Our model achieves SOTA results on several sequential disentanglement benchmarks. While the method relies on heuristics such as initial disentanglement by restricting the dimensions of $s$ and $d_{1:T}$, and the methodology of selecting negative and positive samples, it is backed up with extensive empirical results that show the significance of each component and the robustness of the method to different modalities. Our model uses a similar number of hyperparameters as existing work. Tuning several hyperparameters may be challenging in practice. Nevertheless, we utilized automatic tuning tools, such as hyperopt, to search for the best parameter values within the predefined hyperparameter space. Finally, similar to existing disentanglement works, we used pre-trained classifiers to evaluate our approach. In general, we believe that the sequential disentanglement community will benefit from new challenging benchmarks that depend on improved evaluation metrics.

\section{Discussion}

In this work, we investigated the problem of unsupervised disentanglement of sequential data. Recent SOTA methods employ self-supervision via auxiliary tasks and DA which unfortunately, are modality-based. Namely, they depend on the domain-modality (e.g., videos), on the task-modality (e.g., classifying expressions), or on both. In contrast, we propose a contrastive estimation framework that is free of external signals, and thus is applicable to arbitrary sequential data and tasks. Key to our approach is the observation that VAEs naturally support the generation, comparison, and sampling of distributions. Therefore, effective sampling strategies for generating positive and negative views can be devised based solely on the batch inputs. Our method is easy to code, efficient, and it uniformly treats similar and dissimilar views. Our extensive evaluation shows new SOTA results on multiple datasets including video, audio and arbitrary time series and on downstream tasks as speaker verification, unconditional generation, and prediction.

In the future, we would like to explore the interplay between the mutual information loss components. Essentially, these terms are contradicting in nature, and thus, it motivates us to find improved formulations. Moreover, we would like to investigate whether sampling strategies as our method can be effective for non-sequential contrastive estimation on e.g., static images. We believe that this is a very interesting direction for future research and that with some adaptions, our method can contribute to contrastive learning of static information as well. Finally, we aim to tackle challenging datasets as Jesters using improved VAE pipelines.

\section*{Acknowledgements}

This research was partially supported by the Lynn and William Frankel Center of the Computer Science Department, Ben-Gurion University of the Negev, an ISF grant 668/21, an ISF equipment grant, and by the Israeli Council for Higher Education (CHE) via the Data Science Research Center, Ben-Gurion University of the Negev, Israel.

\clearpage
\bibliography{refs}
\bibliographystyle{icml2023}

% APPENDIX
%%%%%%%%%%%%%%%%%%%%%%%%%%%%%%%%%%%%%%%%%%%%%%%%%%%%%%%%%%%%%%%%%%%%%%%%%%%%%%%
%%%%%%%%%%%%%%%%%%%%%%%%%%%%%%%%%%%%%%%%%%%%%%%%%%%%%%%%%%%%%%%%%%%%%%%%%%%%%%%
\clearpage

\appendix
\onecolumn

\section{Experimental Setup}
\label{app:exp_setup}

\subsection{Datasets} 

\paragraph{Sprites.} A dataset introduced by \cite{reed2015deep} that includes animated cartoon characters that have both static and dynamic attributes. The static attributes include variations in skin, tops, pants, and hair color, each of which has six possible options. The dynamic attributes consist of three different types of motion (walking, casting spells, and slashing) that can be performed in three different orientations (left, right, and forward). In total, there are $1296$ unique characters that can perform nine different motions. Each sequence in the dataset consists of eight RGB images with a size of $64 \times 64$ pixels. In our experiments, we use $9000$ samples for training and $2664$ samples for testing.

\paragraph{MUG.} A Facial expression dataset created by \cite{5617662} that includes image sequences of $52$ subjects displaying six different facial expressions (anger, fear, disgust, happiness, sadness, and surprise). Each video in the dataset consists of between $50$ and $160$ frames. In order to create sequences of length 15, as was done in previous work \cite{bai2021contrastively}, we randomly select 15 frames from the original sequences. We then use Haar Cascades face detection to crop the faces and resize them to $64 \times 64$ pixels, resulting in sequences $x \in \mathbb{R}^{15 \times 3 \times 64 \times 64}$. The final dataset consists of $3429$ samples.

\paragraph{TIMIT.} A dataset introduced by \cite{timit} which consists of read speech that is used for acoustic-phonetic research and other speech tasks. It contains $6300$ utterances ($5.4$ hours of audio). There are $10$ sentences per speaker, for a total of $630$ speakers. The dataset includes adult men and women. For the data pre-processing, we follow the same procedure as in prior work~\cite{yingzhen2018disentangled}. We extract spectrogram features ($10$ms frame shift) from the audio, and we sample segments of $200$ms duration ($20$ frames) from the audio, which are used as independent samples.

\paragraph{Jester.} A dataset introduced by \cite{materzynska2019jester}. The Jester dataset comprises of $\num{148,092}$ labeled video segments of more then $1300$ unique individuals making $27$ simple, predefined hand gestures in front of a laptop camera or webcam. The gestures are labeled whereas the subject is not, and thus the dataset contains only dynamic labels. This dataset is significantly more complex in comparison to MUG since there are variations in the background, light, and pose, and more elements in the image are much bigger. We used five gestures (Pushing Hand Away, Rolling Hand Forward, Shaking Hand, Sliding Two Fingers Left, Sliding Two Fingers Right). We extracted videos with $10$ frames where the gap between two frames has been selected by taking the total sequence length divided by $10$. 

\paragraph{Letters.} The Letters dataset \cite{letters} comprises of English letters and numbers written by $66$ individuals, both offline and online handwritten letters. In our setup, we utilized only small English letters (a-z) from the offline subset of the dataset. We created a lexicon of $100$ words, each consisting of seven letters, and then we generated word sequences using images of the letters. As an example, a sequence may appear as "science". We excluded subject number '$61$' due to missing data and generated 100 word sequences using the handwriting of the remaining $65$ subjects. Each subject has two samples for each letter, which were randomly selected. 

\paragraph{Physionet.} The Physionet ICU Dataset \cite{goldberger2000physiobank} is a medical time series corpus of $\num{12,000}$ adult patients' stays in the Intensive Care Unit (ICU). The data includes time-dependent measurements such as physiological signals and lab measurements as well as general information about the patients, such as their age, the reason for their ICU admission, and etc. Additionally, the dataset includes labels that indicate in-hospital mortality. For pre-processing we follow \cite{tonekaboni2022decoupling}.

\paragraph{Air Quality.} The UCI Beijing Multi-site Air Quality dataset \cite{zhang2017cautionary} was collected over four years from March 1st, 2013 to February 28th, 2017. It includes hourly measurements of multiple air pollutants from $12$ nationally controlled monitoring sites. The meteorological data in each air-quality site are matched with the nearest weather station from the China Meteorological Administration. For our experiments we follow \cite{tonekaboni2022decoupling} and pre-process the data by dividing it into samples from different stations and of different months of the year.

\subsection{Disentanglement Metrics}
\paragraph{Accuracy (Acc).} This is a metric that measures the ability of a model to preserve fixed features while generating others. For instance, freeze the dynamic features and sample the static features. The metric computed by using a pre-trained classifier (called $C$ or the "judge"). The classifier training has been on the same train set of the model and testing is on the same test set as of the model. For example, for the MUG dataset, the classifier would output the facial expression and check that it did not change during the static features sampling.

\paragraph{Inception Score (IS).} This is a metric for the generator performance. First, we apply the judge on all the generated sequences $x_{1:T}$. Thus, getting $\dist{p}{y}{x_{1:T}}$ which is the conditional predicted label distribution. Second, we take $p(y)$ which is the marginal predicted label distribution and we calculate the KL-divergence $\kldiv{\dist{p}{y}{x_{1:T}}}{p(y)}$. Finally, we compute $\text{IS} = \exp(\mathbb{E}_x \kldiv{\dist{p}{y}{x_{1:T}}}{p(y)})$.

\paragraph{Inter-Entropy ($H(y|x)$).} This metric reflects the confidence of the classifier $C$ regarding label prediction. Low Intra Entropy means high confidence. We measure it by entering $k$  generated sequences into the classifier and computing $\frac{1}{k} \sum_{i=1}^k H(p(y | x^i_{1:T}))$. 

\paragraph{Intra-Entroy ($H(y)$).} This metric reflects the diversity among the generate sequence. High Intra-Entropy score means high diversity. It is computed by taking the generated sample from the learned prior distribution $p(y)$ and then using the judge output on the predicted labels $y$.

\paragraph{Equal Error Rate (EER).} This metric is used in the TIMIT dataset for speaker verification task evaluation. It measures the value of the false negative rate, or equally, the value of the false positive rate of a model over the speaker verification task. EER is measured when the above rates are equal.

\paragraph{Latent Accuracy (L-Acc).} This metric measures the ability of a model to generate meaningful latent features for a downstream classification task. For instance, for the MUG dataset, taking the static latent factor $s$ of a sample $x$ and trying to predict the subject label or the facial expression label. In such case, a meaningful and disentangled model will produce static features that will contain information about the subject label but not on the facial expression label of $x$. We compute the prediction accuracy by training a Support Vector Machine Classifier for the static and dynamic features. We flatten the dynamic features $d_{1:T}$ into one vector $d$, that is, assuming $d_i \in \mathbb{R}^k$ and $i=1, ..., T$. Then the dimension of the flattened vector $d$ is $k \times T$. The exact dimension changes between dataset types. We split the test set data into two parts (80-20) and use its first part to train the different classifiers and the second one to evaluate the prediction accuracy. Finally, we also train a Random Forest Classifier and KNN Classifier to show the robustness of the benchmark to the classifier choice.

\begin{table*}[ht]
\centering
\caption{Image Model Architecture.}

\footnotesize
\label{tab:arch_img}
\begin{tabular}{ccc} 
    \toprule
    & Encoder &\\
    \midrule
    &Input: $64 \times64 \times 3$ image \\
    & Conv2D$(3,32,4,2,1)$ \\
    & Conv2D$(32,64,4,2,1)$ \\
    & Conv2D$(64,128, 4,2,1)$ \\
    & Conv2D$(128, 256, 4,2,1)$ \\
    & Conv2D$(256, 128, 4,2,1)$ \\
    & Where each Conv2D layer is followed by BN2D and LeakyReLU activation \\
    & Bidi-LSTM$(128, 256)$\\
    & $s^\mu = \text{Linear}(512, d_s)$, $s^{\text{log}(\sigma)} = \text{Linear}(512, d_s)$\\
    & RNN$(512, 256)$ \\
        &$d_{1:T}^\mu = \text{Linear}(256, d_d)$, $d_{1:T}^{\text{log}(\sigma)} = \text{Linear}(256, d_d)$\\
    \midrule
    & Decoder \\
    \midrule
    & Reparameterize to obtain $s$ and $d_{1:T}$ \\
    & concat $(s,d_{1:T}) = z$ with $d_z$ dimension size \\
    & Conv2DT$(d_z, 256, 4, 1, 0)$ \\
    & Conv2DT$(256, 128, 4, 1, 0)$ \\
    & Conv2DT$(128, 64, 4, 1, 0)$ \\
    & Conv2DT$(64, 32, 4, 1, 0)$ \\
    & Where each Conv2DT layer is followed by BN2D and LeakyReLU activation \\
    & Output: Sigmoid(Conv2DT$(32, 3, 4, 1, 0)$) \\
    \bottomrule
\end{tabular}
\end{table*}

\subsection{Model Architecture}
\label{app:model_arch}
All the models have been implemented using Pytorch~\cite{NEURIPS2019_9015}. The Conv2D and Conv2DT denote a 2D convolution layer and its transpose, and BN2D is a 2D batch normalization layer. 

\paragraph{Image Datasets.} Our image model architecture follows \cite{zhu2020s3vae} implementation. The static latent distribution variables $s^\mu , s^{\text{log}(\sigma)}$ are parameterized by taking the last hidden state of a bi-directional LSTM and propagating it through linear layers. The dynamic latent distribution variables $d_{1:T}^\mu\ , d_{1:T}^{\text{log}(\sigma)}$ are given by propagating the hidden states of the bi-directional LSTM through a RNN and then Linear layers. In Tab.~\ref{tab:arch_img} we describe the encoder and the decoder of our model. Sprites, MUG, Letters, and Jesters, share the same architecture. We denote the dimension of $d_{1:T}$ by $d_d$, and $s$ dimension as $d_s$. The values are chosen per dataset and reported in Tab.~\ref{tab:hypers}.

\paragraph{Audio Datasets.} The architecture of the TIMIT dataset model follows \cite{yingzhen2018disentangled} and was used by the previous methods \cite{zhu2020s3vae, bai2021contrastively}. The only difference from the image architecture is the removal of the convolutions from the encoder and the replacement of the decoder with two linear layers. The first linear layer input dimension is $d_z$ and its output dimension is $256$ followed by LeakyReLU activation. Finally, we feed the second linear layer followed by LeakyReLU activation and its output dimension is $200$.

\paragraph{Time Series Datasets.}
The architecture of the time series dataset is simpler. The encoder is composed of $3$ linear layers, $\text{Linear}(10,32) \rightarrow \text{Linear}(32, 64) \rightarrow \text{Linear}(64,32)$ with ReLU activations after each linear layer followed by similar architecture from image models (Bidi-LSTM etc.) to model $s^{\mu}$, $s^{\log(\sigma)}$, $d_{1:T}^{\mu}$, $d_{1:T}^{\log(\sigma)}$. The decoder is composed of a Linear layer that projects the latent codes onto a dimension of size $32$, followed by tanh activation. Then, the output is propagated through an LSTM with a hidden size of $32$. We feed the output of the LSTM to $2$ linear layers, each followed by a ReLU activation, $\text{Linear}(32,64) \text{ and } \text{Linear}(64,32)$. Finally, we project the output onto $2$ linear layers to produce the mean and covariance from which we sample the final output. This architecture follows \cite{tonekaboni2022decoupling}.

\label{app:setup_arch}
\subsection{Hyperparameters}
We estimate the following objective function:
\begin{align} \begin{split} \label{eq:our_final_loss}
    \max_{p, q} & \; \mathbb{E}_{x_{1:T}\sim p_\mathcal{D}}\lambda_1 \mathbb{E}_{\dist{q}{z}{x_{1:T}}} \log \dist{p}{x_{1:T}}{z}\\ &- \lambda_2 \kldiv{\dist{q}{s}{x_{1:T}}}{p(s)} - \lambda_3 \kldiv{\dist{q}{d_{1:T}}{x_{1:T}}}{p(d_{1:T})}\\
    &+ \lambda_4 I_q(d_{1:T}; x_{1:T}) + \lambda_4 I_q(s; x_{1:T}) - \lambda_5 I_q(s; d_{1:T})   \
\end{split} \end{align}

To control the contribution of each loss component we add $\lambda_1$ coefficient to the reconstruction loss, $\lambda_2$ for static KL term, $\lambda_3$ for the dynamic KL term, and finally $\lambda_4, \lambda_5$ to the contrastive terms. The hyperparameter $\lambda_1$ is tuned over $\{1, 2.5, 5, 10\}$, we do not divide the MSE loss by the batch size. $\lambda_2$ is tuned over $\{1, 3, 5, 7, 9\}$, and $\lambda_4$ and $\lambda_5$ are tuned over $\{0.1, 0.5, 1, 2.5, 5\}$ while $\lambda_3$ is fixed to $1$. We used Adam optimizer~\cite{kingma2014adam} with the learning rate chosen from $\{0.001, 0.0015, 0.002\}$. The dimensions of the static and dynamic features which were chosen among $\{128, 256\}$ for the static and $\{32, 64\}$ for the dynamic factors. Our optimal hyperparameters for each task and dataset are given in Table.~\ref{tab:hypers}. All the tasks were trained for at most $600$ epochs. 

\begin{table}[ht]
    \caption{Hyperparameters for all datasets, lr and bsz are abbreviations for learning rate and batch size, respectively.}
    \label{tab:hypers}
    
    \centering
    % \footnotesize
    \begin{tabular}{lccccccccc}
        \toprule
        Dataset & $\lambda_1$ & $\lambda_2$ & $\lambda_3$ & $\lambda_4$ & $\lambda_5$ & lr & bsz & $d_s$ & $d_d$ \\
        \midrule
        Sprites     & $10$ & $5$ & $1$ & $5$ & $1$ & $2\mathrm{e}{-3}$ & $100$ & $256$ & $32$\\
        MUG         & $5$  & $9$ & $1$ & $0.5$ & $2.5$ & $15\mathrm{e}{-4}$ & $16$ & $256$ & $64$ \\
        
        Letters     & $2.5$  & $1$ & $1$ & $5$ & $5$ & $2\mathrm{e}{-3}$ & $64$ & $256$ & $32$\\
        Jesters     & $5$  & $1$ & $1$ & $1$ & $1$ & $1\mathrm{e}{-3}$ & $16$ & $256$ & $64$ \\
        
        TIMIT       & $5$ & $1$ & $1$ & $0.5$ & $1$ &  $1\mathrm{e}{-3}$ & $10$ & $256$ & $64$ \\
        Physionet   & $2.5$ & $7$ & $1$ & $0.1$ & $2.5$ & $1\mathrm{e}{-3}$ & $10$ & $12$ & $4$ \\
        Air Quality  & $2.5$ & $5$ & $1$ & $0.1$ & $2.5$ & $1\mathrm{e}{-3}$ & $10$ & $12$ & $4$ \\
        \bottomrule
    \end{tabular}
\end{table}

\label{app:setup_hyp}

\section{More Experiments, Analyses and Information}
\label{app:impel_details}

\subsection{Method Implementation and Pseudocode}

In what follows, we will explain in detail the implementation of our sampling procedure. In addition, we add a pseudocode in Alg.~\ref{alg:spst} which describes the process and shows how our framework can be implemented.

\begin{algorithm}[t]
    \caption{Static predictive sampling trick}
    \label{alg:spst}
    \begin{algorithmic}
       \STATE {\bfseries Input:} Batch of samples $\{ x_{1:T}^j \}_{j=1}^n \sim p_\mathcal{D}$
       \STATE 
        \STATE $\mathcal{L}_\text{iNCE} \leftarrow 0$
        \STATE $\{ s^j \}_{j=1}^n \sim \dist{\tilde{q}}{\{ s^j \}_{j=1}^n }{\{ x_{1:T}^j \}_{j=1}^n}$

        \FOR{$i=1$ {\bfseries to} $n$}
            
            \FOR{$j=1$ {\bfseries to} $n$}
                        
            \STATE $D_{ij} \leftarrow \kldiv{\dist{\tilde{q}}{s^i}{x_{1:T}^i}}{\dist{\tilde{q}}{s^j}{x_{1:T}^j}}$
            
        \ENDFOR
            \STATE $\chi_i\leftarrow \text{argsort}(D_i)$
        \ENDFOR

        \FOR{$i=1$ {\bfseries to} $n$}
            \STATE $\omega \leftarrow \chi_i[:\floor{\frac{n}{3}}], \rho \leftarrow \chi_i[\floor{\frac{2n}{3}}:] $

            \STATE $S^+ \leftarrow \{ \dist{\tilde{q}}{s^\omega}{x_{1:T}^\omega} \}, S^- \leftarrow \{ \dist{\tilde{q}}{s^\rho}{x_{1:T}^\rho} \}$
            \STATE 
            \STATE /* Sample Positive Example */
            \STATE $\tilde{d}_{1:T} \sim p(d_{1:T}), \tilde{s}^+ \sim S^+$
            
            \STATE $x_{1:T}^+ \sim \dist{p}{x_{1:T}^+}{\tilde{s}^+, \tilde{d}_{1:T}}$

            \STATE $s^+ \sim \dist{q}{s^+}{x_{1:T}^+}$

            \STATE 
            \STATE /* Sample Negative Examples */

            \STATE $\{\tilde{s}^-\}_{j=1}^{2n} \sim S^-$

            \STATE $\{x_{1:T}^-\}_{j=1}^{2n} \sim  \dist{p}{\{x_{1:T}^-\}_{j=1}^{2n}}{\{\tilde{s}^-\}_{j=1}^{2n}, \{d_{1:T}\}_{j=1}^{2n}}$
            \STATE $\{s^-\}_{j=1}^{2n} \sim \dist{q}{\{s^-\}_{j=1}^{2n}}{\{x_{1:T}^-\}_{j=1}^{2n}}$
            \STATE

            \STATE
            $\mathcal{L}_\text{iNCE} \leftarrow \mathcal{L}_\text{iNCE} + \log \frac{\phi(s, s^+)}{\phi(s, s^+) + \sum_{j=1}^{2n} \phi(s, s^{-, j})}$
        \ENDFOR
       
       \STATE {\bfseries return} $\mathcal{L}_\text{iNCE} / n$
    \end{algorithmic}
\end{algorithm}

\begin{enumerate}
    \item \textbf{Producing static ($s$) and dynamic ($d_{1:T}$) distributions:} Let $x_{1:T} \sim p_\mathcal{D}$. Using the model architecture, elaborated in the previous section, we can compute the mean and log variance vectors that represent the $s$ and $d_{1:T}$ posterior distributions: $s^{\mu}$, $s^{\log(\sigma)}$, $d_{1:T}^{\mu}$, $d_{1:T}^{\log(\sigma)}$ 

    \item \textbf{Producing Positive and Negative Views:} W.l.o.g we focus on how to produce positive and negative static views. However, it is a similar process for the positive and negative dynamic views. To produce the positive view, we need a vector $s^+$ that represents a positive view of the static factor (potentially from the same class) and vectors $\tilde{d}_{1:T}$ that represent arbitrary dynamics.
    
    The vectors $\tilde{d}_{1:T}$ can be simply sampled from the prior distribution $\tilde{d}_{1:T} \sim p(d_{1:T})$. To produce $s^+$, we compute the pairwise KL divergence distances matrix $D \in \mathbb{R}^{n \times n}$ for the batch as described in the main text in Eq~\ref{eq:kl_dist}.  We then sort the row $D_{i:}$ in ascending order, and we sample positive views from the \emph{first} third of distributions denoted by $S^+$, whereas negative views are sampled from the \emph{last} third denoted by $S^-$. To increase the variability in the views, our predictive sampling trick generates these examples from \emph{the posterior of the sampled prior}. To achieve this, we sample $\tilde{s}^+ \sim S^+$. Then, the positive instance $x_{1:T}^+$ is defined via $x_{1:T}^+ \sim \dist{p}{x_{1:T}^+}{\tilde{s}^+, \tilde{d}_{1:T}}$.
    Finally, the positive static view is obtained by sampling the posterior: $s^+ \sim \dist{q}{s^+}{x_{1:T}^+}$.
    
    A similar proccess is performed to obtain negative views. We first sample $2n$ examples from $\{\tilde{s}^-\}_{j=1}^{2n} \sim S^-$ where $S^-$ obtained from $D$. Next, the negative instances $\{x_{1:T}^-\}_{j=1}^{2n}$ are defined via
    \begin{equation*}
        \{x_{1:T}^-\}_{j=1}^{2n} \sim  \dist{p}{\{x_{1:T}^-\}_{j=1}^{2n}}{\{\tilde{s}^-\}_{j=1}^{2n}, \{d_{1:T}\}_{j=1}^{2n}} \ .
    \end{equation*}

    Finally, the negative static views are obtained by sampling the posterior
    \begin{equation*}
        \{s^-\}_{j=1}^{2n} \sim \dist{q}{\{s^-\}_{j=1}^{2n}}{\{x_{1:T}^-\}_{j=1}^{2n}} \ .
    \end{equation*}

    The computational complexity of the $D_{KL}$ matrix is an important aspect of this stage. In practice, we never construct this matrix. Instead, we exploit the parallel capabilities of PyTorch. Notice that this computation is parallel on the level of the cell, as each matrix cell is independent of the others. Thus, PyTorch can utilize its full parallelization capabilities on this task. In particular, since the computation per cell is constant in time (and memory), the entire computation of  $D_{KL}$ can be made in constant time, if every cell is calculated by a separate compute node. Therefore, the first for loop in Alg. 1 utilizes the full parallel compute capabilities of PyTorch. Similarly, the second for loop in Alg. 1 is re-phrased in a tensorial form such that the for loop is avoided completely.

    \item \textbf{Calculating the Contrastive Loss:} We can compute the contrastive loss ($\mathcal{L}_\text{iNCE}$) given the positive $s^+$ and the negatives $\{{s^-}\}_{j=1}^{2n}$ samples. 
\end{enumerate}

\subsection{A Thirds Similarity Distributions Experiment}
\label{experiments:thirds_exp}

\begin{figure*}[t]
    \centering
    \begin{overpic}[width=.4\linewidth]{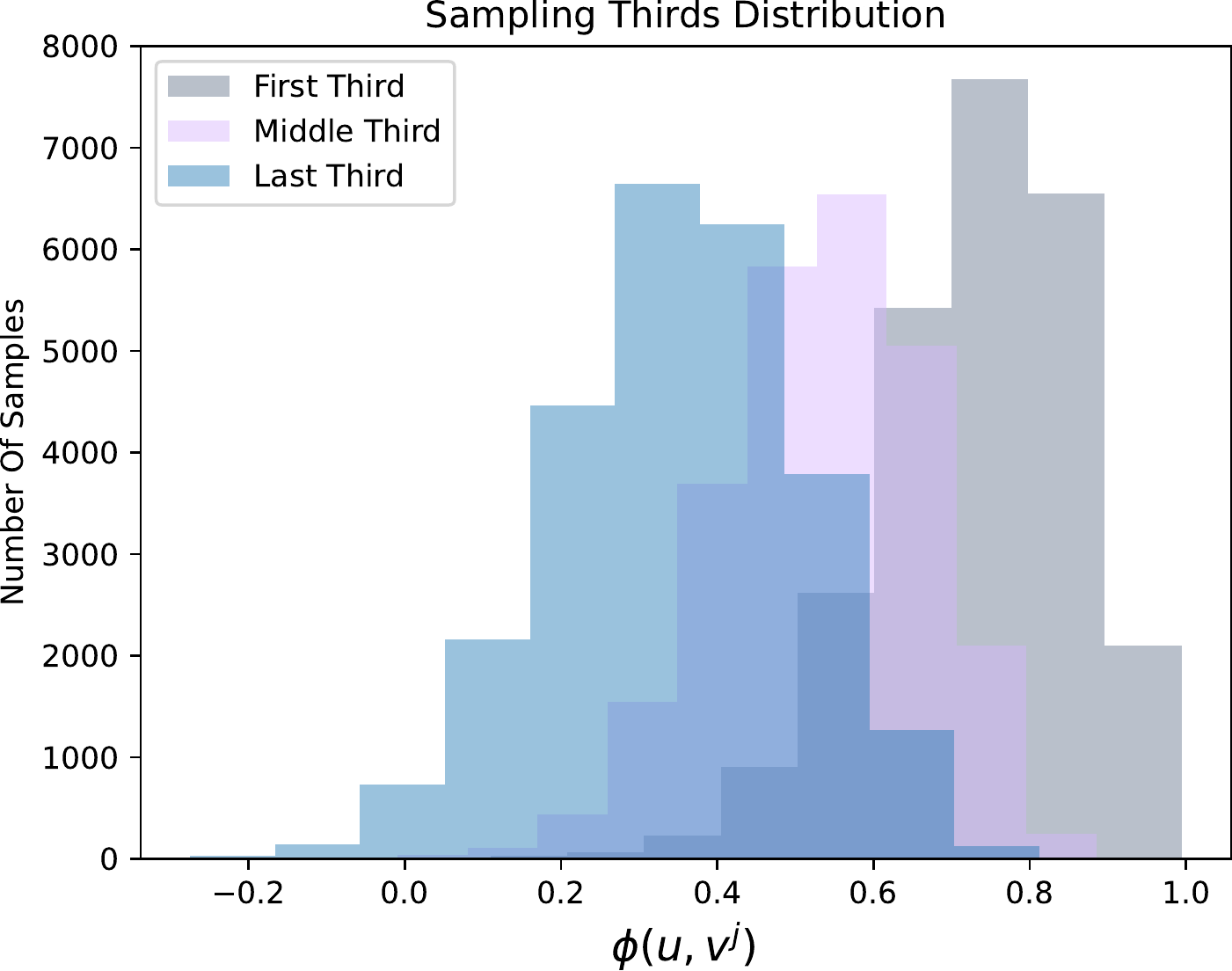}
    \end{overpic}
    \caption{We compute the similarity scores $\phi(u, v^j)$ of the current sample $u$ with respect to samples positioned in the first, middle, and last thirds. The scores are ordered sequentially, i.e., the first third attains the most similar samples, whereas the last third includes the most dissimilar examples.}
    \label{fig:thirds_dist}
\end{figure*}

Throughout our studies, we observed that taking the negatives from the last third, obtained the best results as seen in the ablation study in Sec.~\ref{subsec:ablation}. One possible explanation is that the last third consists of fewer positive points, i.e., samples we consider to be negative but are in fact positive. Since our negative sampling process is random, it might be that using negative samples from the last third avoids positive samples more often than taking such samples from the middle third, which yields better overall results in practice. To strengthen this hypothesis, we calculated the distribution of similarity $\phi(u, v^j)$  for each third. We used our trained model to get the latent space vectors and calculated their similarity scores. We conduct the experiment both for the static and dynamic latent vectors. The results are very similar, thus we decided to show here the dynamic vector similarity distribution. We show the similarity histogram of the various thirds in Fig.~\ref{fig:thirds_dist}. The histogram shows an intuitive ordering of the thirds in the sense that the first third yields the most similar samples, and the last third yields the most dissimilar samples. Thus, we believe that the governing factor which made the last third to be better is that wrong samples are probably sampled less often in comparison to the middle third. In addition, notice that the last third does contain several examples with similarity $\phi(u, v^j)$  higher than $0.5$, and these samples may be hard negative examples.

\subsection{The Predictive Sampling Trick vs a Reparametrization Trick}

To further analyze the contribution of the predictive sampling trick, we re-trained two neural networks with the same hyperparameters on Sprites and MUG without the predictive sampling trick, and instead, we used a simple reparametrization trick. We report the results of the comparison between the two in Tab.~\ref{tab:rep_vs_pred}. One can observe that the reparametrization trick models' results are inferior in comparison to the results of the predictive sampling trick results. For instance, while the accuracy metric for Sprites is saturated, a reduction in the other metrics is noticeable, e.g. $8.942$, using our method vs. $8.865$ using the reparametrization trick on the IS metric. The difference is even more noticeable on the MUG dataset, where the reparametrization trick suffers an almost two percent loss in accuracy. These results further motivate and reinforce our choice and design of the predictive sampling trick.

\setlength{\tabcolsep}{2pt}
\begin{table*}[!t]
    \caption{Disentanglement metrics on Sprites and MUG using only a reparametrization trick (repar. trick) vs. using our predictive sampling trick (ours). Our results are better overall across all metrics.}
    \label{tab:rep_vs_pred}

    \resizebox{\textwidth}{!}{
    \centering
    \footnotesize
        \begin{tabular}[t]{lcccc|cccc}
            \toprule
            & \multicolumn{4}{c|}{Sprites} & \multicolumn{4}{c}{MUG} \\
            Method & Acc$\uparrow$ & IS$\uparrow$ & $H(y|x){\downarrow}$ & $H(y){\uparrow}$ & Acc$\uparrow$ & IS$\uparrow$ & $H(y|x){\downarrow}$ & $H(y){\uparrow}$ \\
            \midrule
            repar. trick & $100\% \pm 0\%$ & $8.865 \pm \num{9.98e-4}$ & $0.015 \pm \num{1.13e-4}$ & $2.197 \pm 0$ & $83.93\% \pm 0.96$ & $5.495 \pm 0.048$ & $0.092 \pm \num{7.9e-3}$ & $1.775 \pm \num{4.2e-3}$ \\
            \midrule
            Ours & $100\% \pm 0\%$ & $\boldsymbol{8.942 \pm \num{3.3e-5}}$ & $\boldsymbol{0.006 \pm \num{4e-6}}$ & $\boldsymbol{2.197 \pm 0}$ & $\boldsymbol{85.71\% } \pm 0.9$ & $\boldsymbol{5.548 } \pm 0.039$ & $\boldsymbol{0.066 \pm \num{4e-3}}$ & $\boldsymbol{1.779 \pm \num{6e-3}}$ \\
            \bottomrule
        \end{tabular} }
\end{table*}

\subsection{Qualitative Evaluations}
\label{app:latent_emb}

\begin{figure*}[t]
    \centering
    \begin{overpic}[width=1\linewidth]{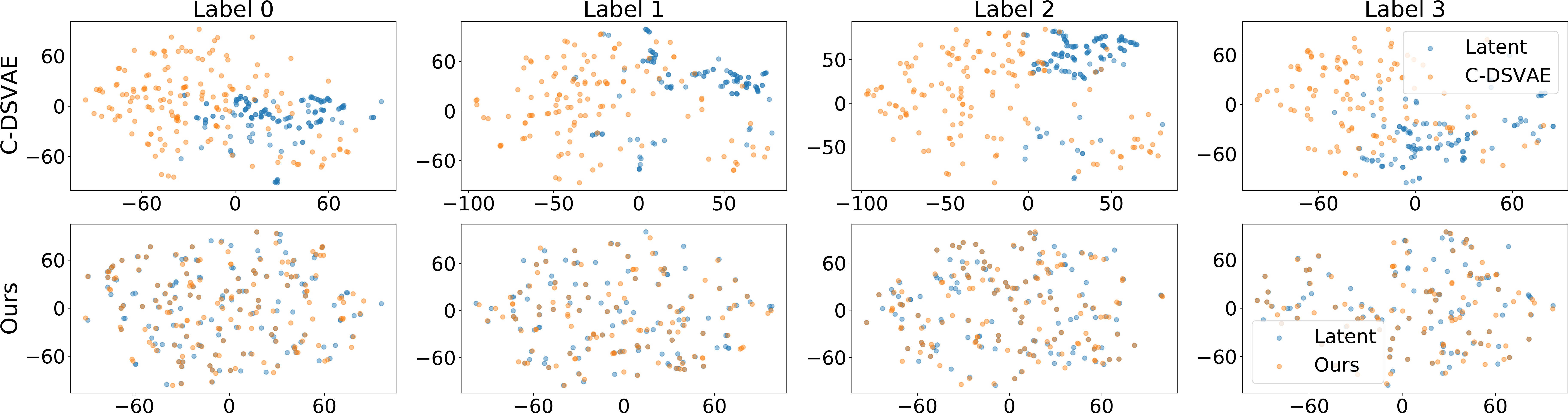}
    \end{overpic}
    \caption{We plot the t-SNE embedding of the dynamic features of the inputs and their positive samples as computed by C-DSVAE (top) and our approach (bottom). In this setting, the blue and orange embeddings should be as close as possible.}
    \label{fig:vis_latent_pos}
\end{figure*}

\begin{figure*}[t]
    \centering
    \begin{overpic}[width=1\linewidth]{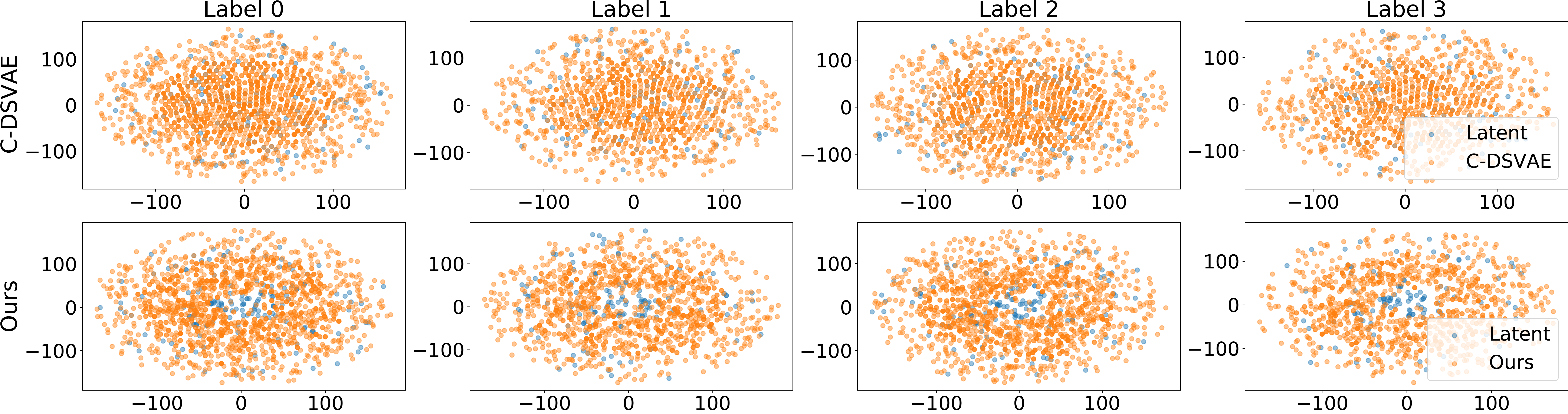}
    \end{overpic}
    \caption{We plot the t-SNE embedding of the dynamic features of the inputs and their negative samples as computed by C-DSVAE (top) and our approach (bottom). In this setting, the blue and orange embeddings should be as separate as possible.}
    \label{fig:vis_latent_neg}
\end{figure*}

Here, we propose an additional qualitative evaluation of our contrastive estimation using the MUG dataset. Specifically, we evaluated the positive and negative samples on two trained models, C-DSVAE and ours. We collect the dynamic latent representations $d_{1:T}^i$ for every sample $i$ in the test set, and we compute the mean value $d = \frac{1}{T}\sum_{j=1}^T d_j$, where the index $i$ is omitted for brevity. For each of those samples, we extract a subset of their positive $d^+$ and negative $d^-$ samples. To visualize these latent features, we project the original representation $d$ and the new samples $d^+$ and $d^-$ using t-SNE~\cite{van2008visualizing}. We anticipate that the pair $(d, d^+)$ will be close in latent space as contrastive learning attracts positive data points closer. In contrast, contrastive learning repels negative samples, and thus $(d, d^-)$ should be far from each other. We present the results in~\ref{fig:vis_latent_pos} and \ref{fig:vis_latent_neg}. For the positive samples, our method shows an impressive similarity between $d$ and $d^+$. In comparison, C-DSVAE presents a much bigger distance between the $d$ and $d^+$. On the negative samples, our method shows good discrimination between $d$ and $d^-$, and in addition, our samples are much more concentrated. In comparison, discriminating between the input and negative samples in C-DSVAE is more challenging. We obtained similar results for the static setting, i.e., when we studied the t-SNE embeddings of $s$ and its positive $s^+$ and negative $s^-$ samples.

\subsection{Additional Information on the TIMIT Speaker Verification Task}
\label{app:timit}

Here, we discuss more on the Audio experiment with TIMIT described in Sec.~\ref{subsection:common_bench}. First, the TIMIT test dataset contains eight different sentences for each speaker with $24$ unique speakers. In total there are $192$ audios. For all those audios, we extract their $s$ and $d_{1:T}$ latent representations. Then, to prepare a single vector representation we calculate the identity representation vector by the same procedure described in \cite{yingzhen2018disentangled}. Last, the EER is being calculated separately for all the combinations of $192$ vectors. In total, there are $\num{18336}$ pairs. We repeat this process independently once for the $s$ features and once for the $d_{1:T}$ features. 

\subsection{Standard Deviation Measures for Tab.~\ref{tab:gen_sprites_mug_timit}}
\label{app:table_1_expand}

Here, we report the standard deviation measures related to Tab.~\ref{tab:gen_sprites_mug_timit} in the main text. Notice that the audio experiment is deterministic due to the ERR metric definition, and thus it does not have standard deviation measures. The extended results are provided in Tab.~\ref{tab:table_1_expanded}. These results indicate that not only our method achieves superior results in comparison to SOTA approaches, but also, it is well within the statistical significance regime, given the standard deviation measures.

\setlength{\tabcolsep}{2pt}
\begin{table*}[!b]
    \caption{We augment Tab.~\ref{tab:gen_sprites_mug_timit} for the Sprites and MUG datasets with standard deviation measures.}
    \label{tab:table_1_expanded}

    \resizebox{\textwidth}{!}{
    \centering
    \footnotesize
        \begin{tabular}[t]{lcccc|cccc}
            \toprule
            & \multicolumn{4}{c|}{Sprites} & \multicolumn{4}{c}{MUG} \\
            Method & Acc$\uparrow$ & IS$\uparrow$ & $H(y|x){\downarrow}$ & $H(y){\uparrow}$ & Acc$\uparrow$ & IS$\uparrow$ & $H(y|x){\downarrow}$ & $H(y){\uparrow}$ \\
            \midrule
            MoCoGAN & $92.89\%$ & $8.461$ & $0.090$ & $2.192$ & $63.12\%$ & $4.332$ & $0.183$ & $1.721$ \\
            DSVAE & $90.73\%$ & $8.384$ & $0.072$ & $2.192$ & $54.29\%$ & $3.608$ & $0.374$ & $1.657$ \\
            R-WAE & $98.98\%$ & $8.516$ & $0.055$ & $\boldsymbol{2.197}$ & $71.25\%$ & $5.149$ & $0.131$ & $1.771$ \\
            S3VAE & $99.49\%$ & $8.637$ & $0.041$ & $\boldsymbol{2.197}$ & $70.51\%$ & $5.136$ & $0.135$ & $1.760$ \\
            SKD & $\boldsymbol{100\%}$ & $\boldsymbol{8.999}$ & $\boldsymbol{\expnumber{1.6}{-7}}$ & $\boldsymbol{2.197}$ & $77.45\%$ & $\boldsymbol{5.569}$ & $\boldsymbol{0.052}$ & $1.769$ \\
            C-DSVAE & $99.99\%$ & $8.871$ & $0.014$ & $\boldsymbol{2.197}$ & $81.16\%$ & $5.341$ & $0.092$ & $1.775$ \\
            \midrule
            Ours & $\boldsymbol{100\% \pm 0\%}$ & $8.942 \pm \num{3.3e-5}$ & $0.006 \pm \num{4e-6}$ & $\boldsymbol{2.197 \pm 0}$ & $\boldsymbol{85.71\% } \pm 0.9$ & $\boldsymbol{5.548 } \pm 0.039$ & $0.066 \pm \num{4e-3}$ & $\boldsymbol{1.779 \pm \num{6e-3}}$ \\
            \bottomrule
        \end{tabular} }
\end{table*}

\subsection{Data Generation}

\begin{figure*}[t]
    \centering
    \begin{overpic}[width=1\linewidth]{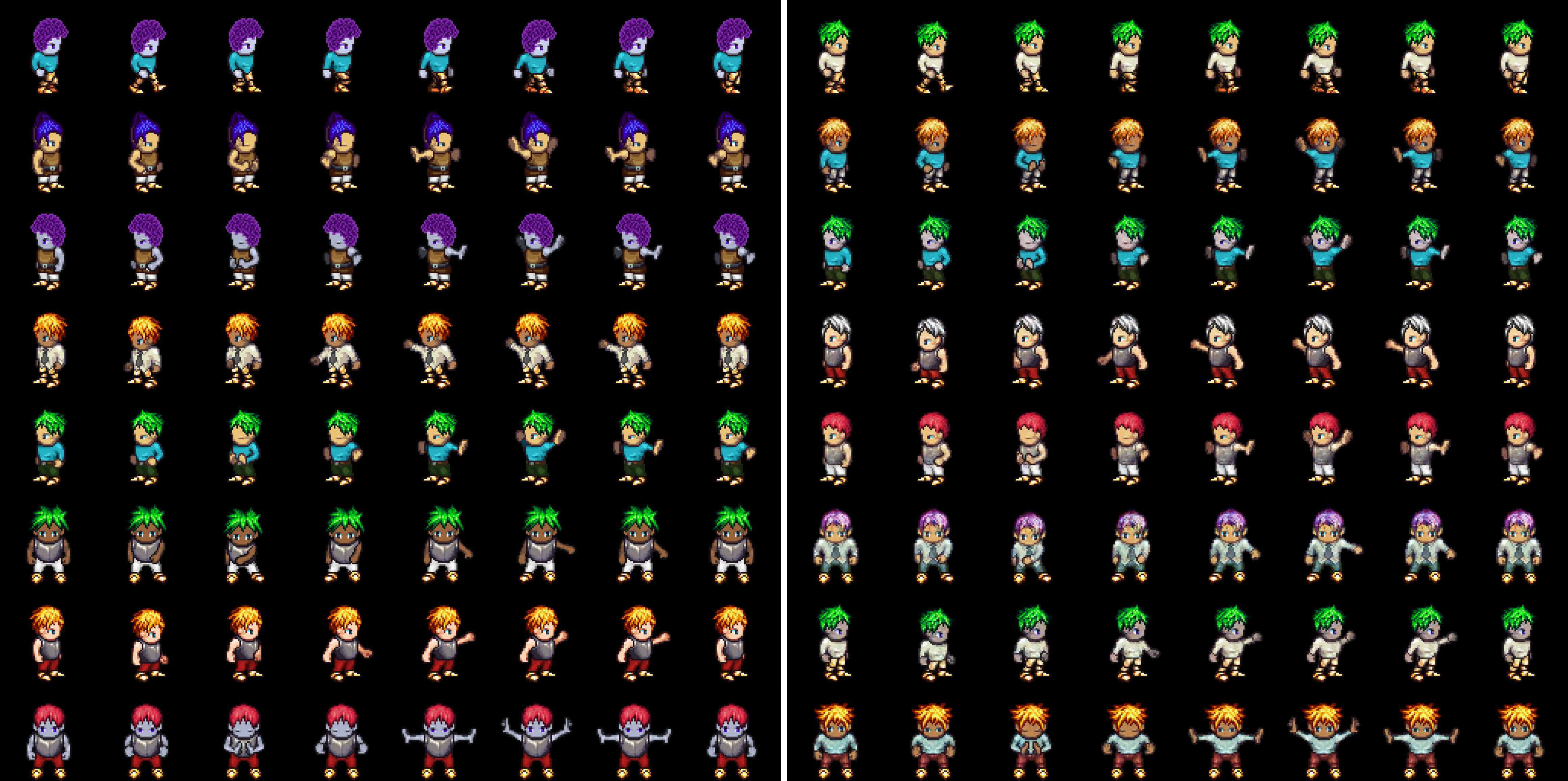}
        % \put(.5, 23){A} \put(3, 12){$s$}
    \end{overpic}
    \caption{Content generation results in Sprites dataset. See the text for additional details.}
    \label{.fig:gen_sprites_content}
\end{figure*}

\begin{figure*}[ht]
    \centering
    \begin{overpic}[width=1\linewidth]{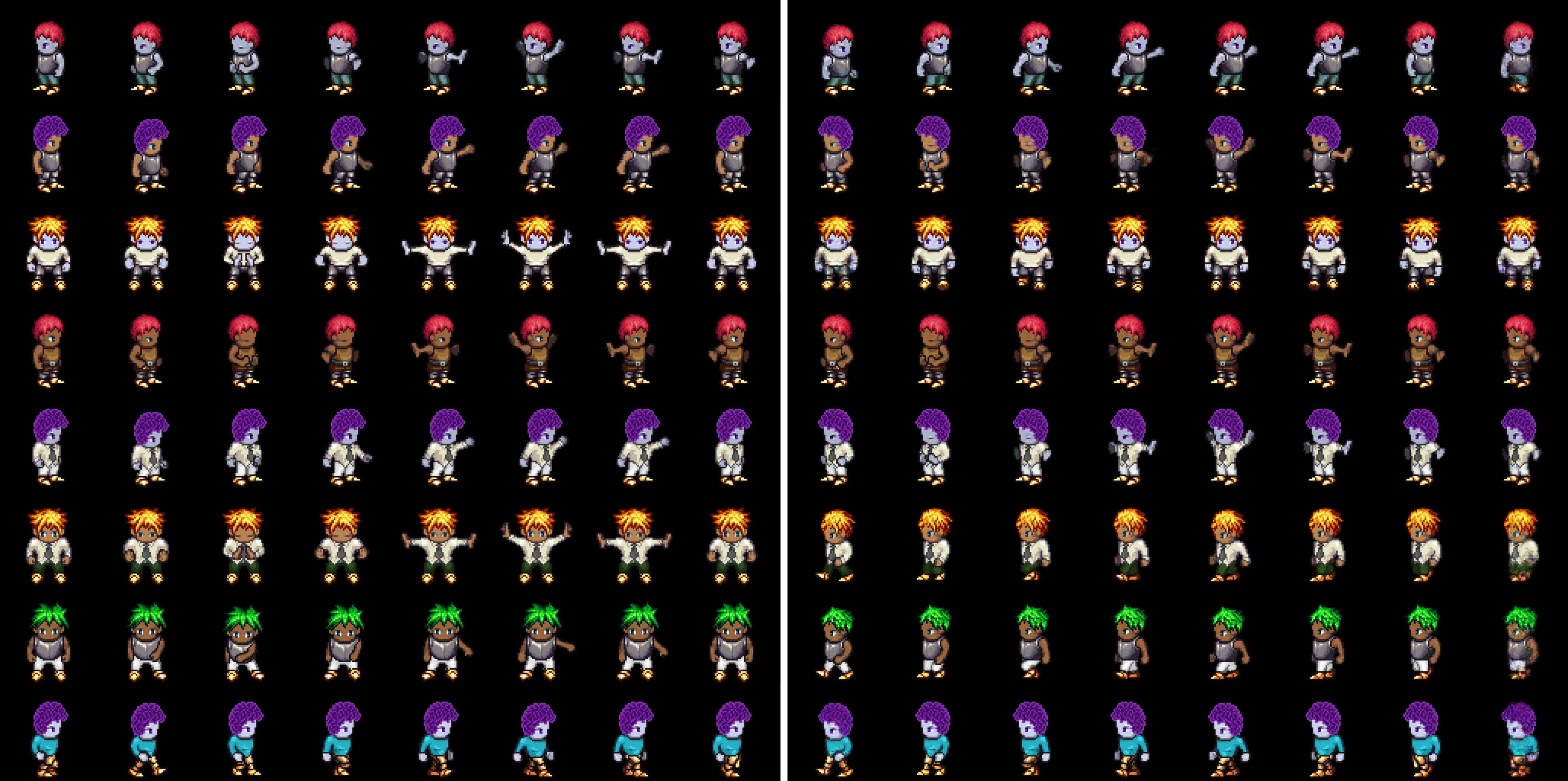}
        % \put(.5, 23){A} \put(3, 12){$s$}
    \end{overpic}
    \caption{Dynamics generation results in Sprites dataset. See the text for additional details.}
    \label{.fig:gen_sprites_dynamics}
\end{figure*}

\begin{figure*}[t]
    \centering
    \begin{overpic}[width=1\linewidth]{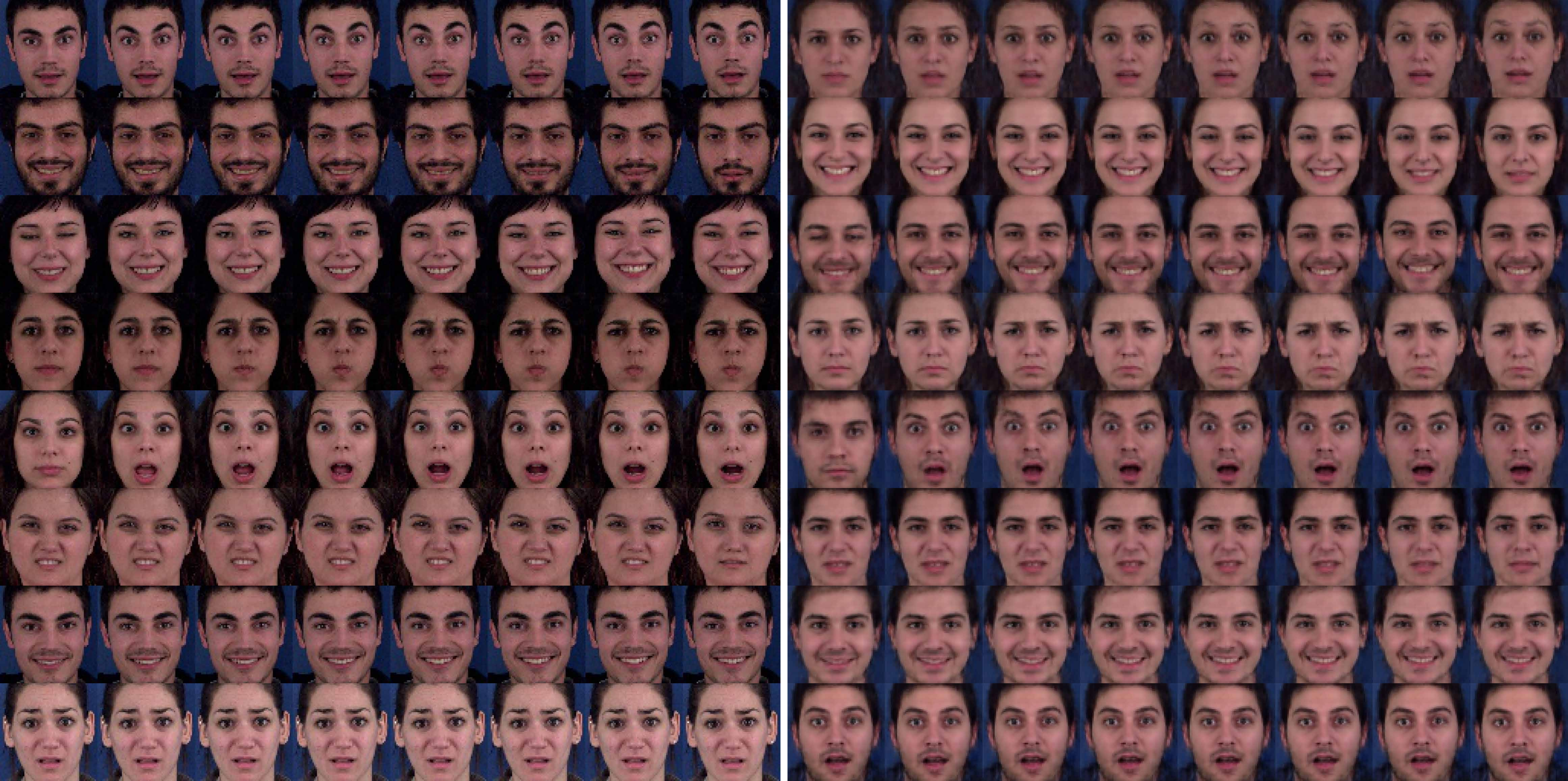}
        % \put(.5, 23){A} \put(3, 12){$s$}
    \end{overpic}
    \caption{Content generation results in MUG dataset. See the text for additional details.}
    \label{.fig:gen_mug_content}
\end{figure*}

\begin{figure*}[ht]
    \centering
    \begin{overpic}[width=1\linewidth]{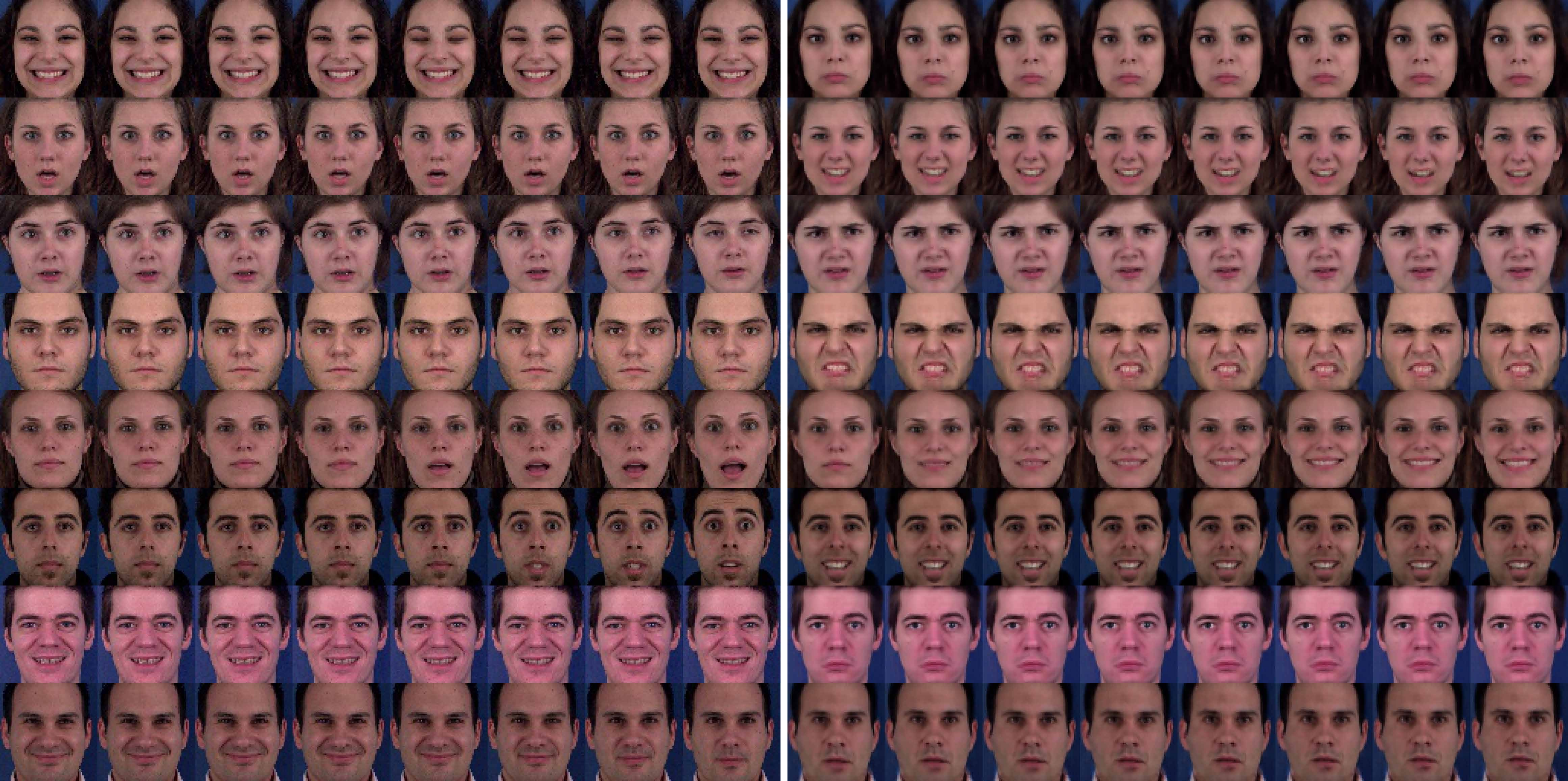}
        % \put(.5, 23){A} \put(3, 12){$s$}
    \end{overpic}
    \caption{Dynamics generation results in MUG dataset. See the text for additional details.}
    \label{.fig:gen_mug_dynamics}
\end{figure*}

We qualitatively evaluate our model's ability to generate static and dynamics features. Specifically, let $x_{1:t} \sim p_D$ denote a sample from the data with its static and dynamic factors latent representations $(s, d_{1:T})$ given by $s \sim \dist{q}{s}{x_{1:T}}$ and $d_{1:T} \sim \dist{q}{d_{1:T}}{x_{1:T}}$. We generate new static features by sampling from the static prior distribution $p(s)$, namely, $\Tilde{s} \sim p(s)$ and fixing the dynamics $d_{1:T}$. Then, we concatenate $(\tilde{s}, d_{1:T})$, and we generate a new sample $\tilde{x}_{1:T} \sim \dist{p}{\tilde{x}_{1:T}}{\tilde{s}, d_{1:T}}$. Finally, we perform a similar process in order to generate the dynamics, where we sample from the dynamic prior distribution and the static features are fixed. The results of static and dynamic features' generation for the Sprites and MUG datasets are given in Fig.~\ref{.fig:gen_sprites_content}, Fig.~\ref{.fig:gen_sprites_dynamics}, Fig.~\ref{.fig:gen_mug_content}, and Fig.~\ref{.fig:gen_mug_dynamics}. The left column in each figure contains the original samples and the right column contains the generated samples. If the model disentangles the features well and has high generation performance, then, the fixed features should be preserved perfectly and the generated features should be random (independent of the original class).

\subsection{Swaps}
\begin{figure*}[t]
    \centering
    \begin{overpic}[width=1\linewidth]{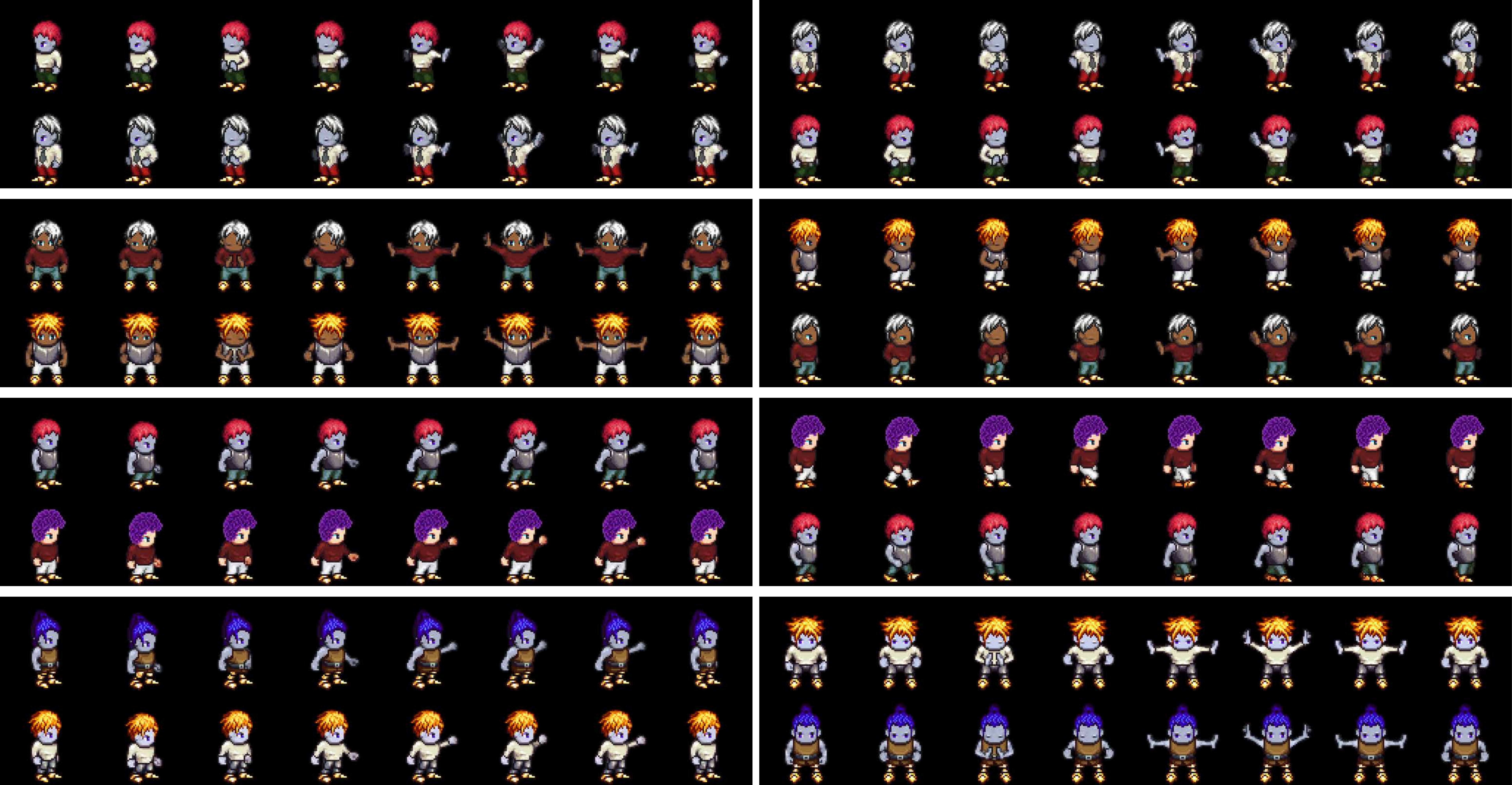}
        % \put(.5, 23){A} \put(3, 12){$s$}
    \end{overpic}
    \caption{Swapping results in Sprites dataset. See the text for additional details.}
    \label{.fig:sprites_swap_app}
\end{figure*}

\begin{figure*}[ht]
    \centering
    \begin{overpic}[width=1\linewidth]{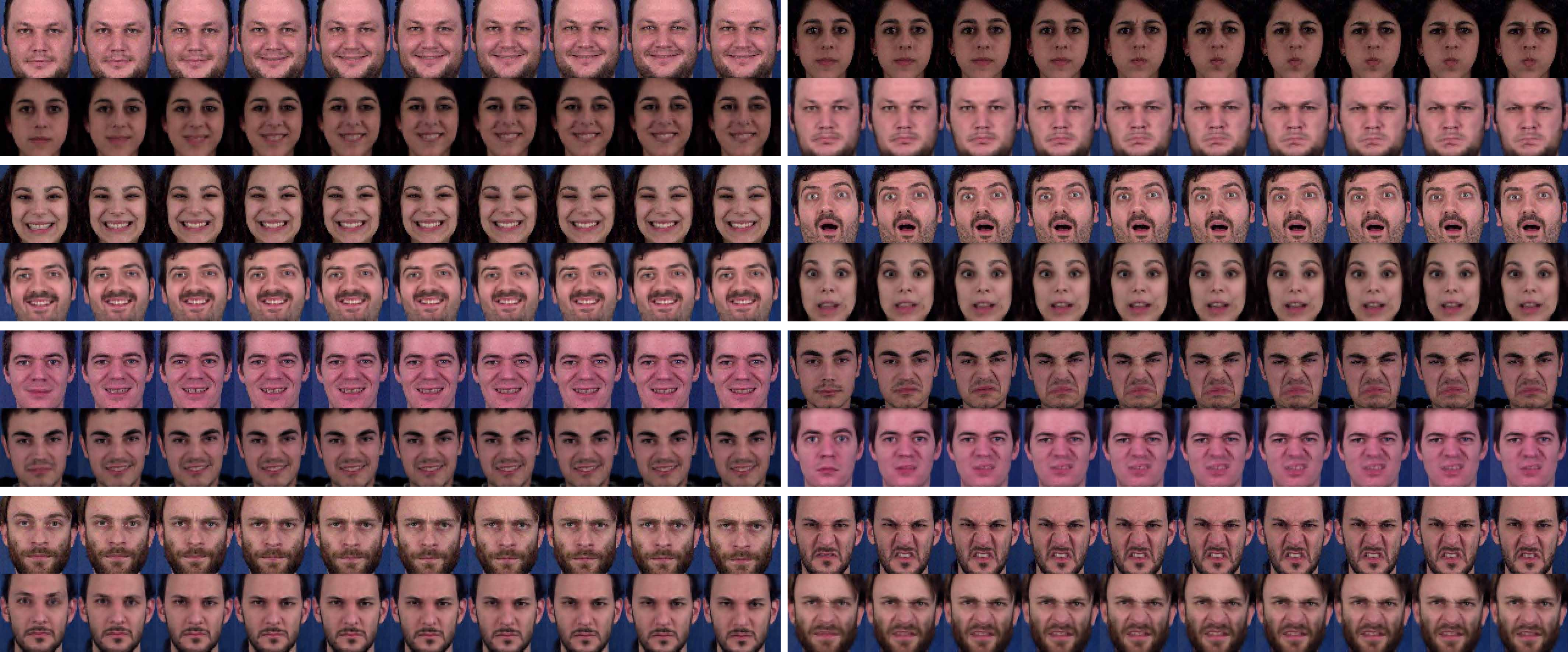}
        % \put(.5, 23){A} \put(3, 12){$s$}
    \end{overpic}
    \caption{Swapping results in MUG dataset. See the text for additional details.}
    \label{.fig:mug_swap_app}
\end{figure*}

In this section we perform another qualitative experiment. Specifically, given source and target samples, $x_{1:T}^\text{src}, x_{1:T}^\text{tgt} \sim p_D$, we swap the static and dynamic features between the source
and the target. In practice, we feed the encoder with the samples to extract their static and dynamic latent representation, $(s^\text{src},d_{1:T}^\text{src})$ s.t $s^\text{src} \sim \dist{q}{s^\text{src}}{x_{1:T}^\text{src}}$ and $d_{1:T}^\text{src} \sim \dist{q}{d_{1:T}^\text{src}}{x_{1:T}^\text{src}}$ for the source and $s^\text{tgt} \sim \dist{q}{s^\text{tgt}}{x_{1:T}}^\text{tgt}$ and $d_{1:T}^\text{tgt} \sim \dist{q}{d_{1:T}^\text{tgt}}{x_{1:T}^\text{tgt}}$ for the target. Then, we generate swapped samples by feeding the decoder, $\Tilde{x}_{1:T}^\text{src} \sim \dist{p}{\Tilde{x}_{1:T}^\text{src}}{\Tilde{s}^\text{tgt}, d_{1:T}^\text{src}}$
and 
$\Tilde{x}_{1:T}^\text{tgt} \sim \dist{p}{\Tilde{x}_{1:T}^\text{tgt}}{\Tilde{s}^\text{src}, d_{1:T}^\text{tgt}}$. If the representation is well disentangled, $\Tilde{x}_{1:T}^\text{src}$ should preserve its original dynamics but have the target's static features and vice versa for $\Tilde{x}_{1:T}^\text{tgt}$. Fig.~\ref{.fig:sprites_swap_app} and Fig.~\ref{.fig:mug_swap_app} show four separate examples of Sprites and MUG where the length of the MUG sequences are shorten to $T=10$ for clarity. The first row of each pair shows the original samples $x_{1:T}^\text{src}, x_{1:T}^\text{tgt} \sim p_D$. The row below shows the swapping results. Namely, rows $1, 3, 5 ,7$ represent the original samples and rows $2, 4, 6, 8$ represent swapped samples.

\subsection{Latent Classification Experiments with Different Classifiers and Standard Deviation}
\label{app:ltnt_features_task}

Here, we elaborate more on the experiment we reported in Sec.~\ref{exp:latent_features_exp}. We extracted the static $s$ and dynamic $d_{1:T}$ features using a trained model and trained four different classifiers. All trained classifiers are Support Vector Machines. We used the default Support Vector Classifier (SVC) of the sklearn package without changing any hyperparameter. To strengthen the statistical significance of Tab.~\ref{tab:dis_met_cls_task} from the main text, we conduct the same experiment with different classifiers and report their results in Tab.~\ref{tab:dis_met_cls_task_svc} (SVC), Tab.~\ref{tab:dis_met_cls_task_random_forest} (Random Forest Classifier), and Tab.~\ref{tab:dis_met_cls_task_knn} (KNN). These tables show that our results are robust to the choice of the classifier. We repeated the experiments per classifier for $10$ times with different seeds for data splitting and report their means and standard deviations. We used the default sklearn Random Forest Classifier and KNN. We used the sklearn default hyperparameters and conducted the same experiment procedure exactly. In the Jesters dataset, we do the exact same process just without the static features and their classifiers since the subjects in this data are not labeled. In the Letters dataset, there is one difference. For each $d_{i}$ , $i=1,...,T$ in $d_{1:T}$ we try to predict its corresponding letter label in the sequence instead of trying to predict the whole word. Briefly, our model maintains its superior performance in comparison to C-DSVAE~\cite{bai2021contrastively} with respect to the gap metric among all classifiers.

\begin{table*}
    \caption{Downstream classification task on latent static and dynamic features \textbf{using SVC}.}
    \label{tab:dis_met_cls_task_svc}
    \centering
    \footnotesize

    \begin{tabular}[t]{ll|ccc|ccc}
        \toprule
        & & \multicolumn{3}{c|}{Static features} & \multicolumn{3}{c}{Dynamic features} \\
        Dataset & Method  & Static L-Acc $\uparrow$ & Dynamic L-Acc $\downarrow$ & Gap $\uparrow$ & Static L-Acc  $\downarrow$ & Dynamic L-Acc $\uparrow$ & Gap $\uparrow$ \\
        \midrule
        \multirow{3}{*}{MUG} & random & $1.92 \%$ & $16.66\%$ & - & $1.92 \%$ & $16.66\%$ & - \\
        & C-DSVAE    & $\boldsymbol{98.75\% \pm 1\%} $ & $76.25\% \pm 2.9\%$ & $22.25\%$ & $26.25\% \pm 3.8\%$ & $82.50\% \pm 2.5\%$ & $56.25\%$ \\
        & Ours       & $98.12\% \pm 0.9\%$ & $\boldsymbol{68.75\% \pm 2.6\%}$ & $\boldsymbol{29.37\%}$ & $\boldsymbol{10.00\% \pm 3.4\%}$ & $\boldsymbol{85.62\% \pm 2.4\%}$ & $\boldsymbol{75.62\%}$ \\
        \midrule
        \multirow{3}{*}{Letters} & random  & $1.65 \%$ & $3.84 \%$ & - & $1.65 \%$ & $3.84 \%$ & - \\
        & C-DSVAE    & $95.47\% \pm 0.5\%$ & $13.0\% \pm 0.6\%$ & $82.47\%$ & $\boldsymbol{2.79\% \pm 0.5\%}$ & $66.35\% \pm 1\%$ & $63.56\%$ \\
        & Ours       & $\boldsymbol{100\% \pm 0\%} $ & $\boldsymbol{12.16\% \pm 0.6\%}$ & $\boldsymbol{87.84\%}$ & $3.06\% \pm 0.3\%$ & $\boldsymbol{69.75\% \pm 1.4\%}$ & $\boldsymbol{66.69\%}$ \\
        \midrule
    \end{tabular}
\end{table*}

\begin{table*}
    \caption{Downstream classification task on latent static and dynamic features \textbf{using Random Forest Classifier}.}
    \label{tab:dis_met_cls_task_random_forest}
    \centering
    \footnotesize
    \begin{tabular}[t]{ll|ccc|ccc}
        \toprule
        & & \multicolumn{3}{c|}{Static features} & \multicolumn{3}{c}{Dynamic features} \\
        Dataset & Method  & Static L-Acc $\uparrow$ & Dynamic L-Acc $\downarrow$ & Gap $\uparrow$ & Static L-Acc  $\downarrow$ & Dynamic L-Acc $\uparrow$ & Gap $\uparrow$ \\
        \midrule
        \multirow{3}{*}{MUG} & random & $1.92 \%$ & $16.66\%$ & - & $1.92 \%$ & $16.66\%$ & - \\
        & C-DSVAE    & $\boldsymbol{98.75\% \pm 0.7\%} $ & $72.75\% \pm 2.8\%$ & $26\% $ & $64\% \pm 4.1\%$ & $83.18\% \pm 1.7\%$ & $19.18\%$ \\
        & Ours       & $98.06\% \pm 1.1\%$ & $\boldsymbol{68.50\% \pm 2.7\%}$ & $\boldsymbol{29.56\%}$ & $\boldsymbol{38.93\% \pm 4.6\%}$ & $\boldsymbol{86.75\% \pm 2.5\%}$ & $\boldsymbol{47.82\%}$ \\
        \midrule
        \multirow{3}{*}{Letters} & random  & $1.65 \%$ & $3.84 \%$ & - & $1.65 \%$ & $3.84 \%$ & - \\
        & C-DSVAE    & $100\% \pm 0\%$ & $5.6\% \pm 0.6\%$ & $94.4\%$ & $3.66\% \pm 0.3\%$ & $75.44\% \pm 1\%$ & $71.78\%$ \\
        & Ours       & $100\% \pm 0\%$ & $5.6\% \pm 0.5\%$ & $94.4\%$ & $\boldsymbol{2.92\% \pm 0.3\%}$ & $\boldsymbol{77.63\% \pm 1\%}$ & $\boldsymbol{74.71\%}$ \\
        \midrule
    \end{tabular}
\end{table*}

\begin{table*}

    \caption{Downstream classification task on latent static and dynamic features \textbf{using KNN}.}
    \label{tab:dis_met_cls_task_knn}
    \centering
    \footnotesize
    \begin{tabular}[t]{ll|ccc|ccc}
        \toprule
        & & \multicolumn{3}{c|}{Static features} & \multicolumn{3}{c}{Dynamic features} \\
        Dataset & Method  & Static L-Acc $\uparrow$ & Dynamic L-Acc $\downarrow$ & Gap $\uparrow$ & Static L-Acc  $\downarrow$ & Dynamic L-Acc $\uparrow$ & Gap $\uparrow$ \\
        \midrule
        \multirow{3}{*}{MUG} & random & $1.92 \%$ & $16.66\%$ & - & $1.92 \%$ & $16.66\%$ & - \\
        & C-DSVAE    & $98.31\% \pm 1.1\% $ & $50.31\% \pm 3.6\%$ & $48\% $ & $26.18\% \pm 3.4\%$ & $\boldsymbol{83.25\% \pm 2.1\%}$ & $57.07\%$ \\
        & Ours       & $\boldsymbol{99.31\% \pm 0.9\%} $ & $\boldsymbol{49.25\% \pm 4.5\%}$ & $\boldsymbol{50.06\%}$ & $\boldsymbol{19.31\% \pm 2.8\%}$ & $82.50\% \pm 1.5\%$ & $\boldsymbol{63.19\%}$ \\
        \midrule
        \multirow{3}{*}{Letters} & random  & $1.65 \%$ & $3.84 \%$ & - & $1.65 \%$ & $3.84 \%$ & - \\
        & C-DSVAE    & $100\% \pm 0\%$ & $5.8\% \pm 0.5\%$ & $94.2\%$ & $\boldsymbol{3.41\% \pm 0.3\%}$ & $76.61\% \pm 0.9\%$ & $73.2\%$ \\
        & Ours       & $100\% \pm 0\%$ & $5.8\% \pm 0.6\%$ & $94.2\%$ & $3.54\% \pm 0.3\%$ & $\boldsymbol{78.33\% \pm 0.8\%}$ & $\boldsymbol{74.92\%}$  \\
        \midrule
    \end{tabular}
\end{table*}

\subsection{Robustness with Respect to the Seed Choice}

In our work, we based our evaluation section with respect to existing state-of-the-art models and their evaluation protocol and benchmark datasets. Following these approaches, the sensitivity to hyperparameters and randomness is typically not considered. Nevertheless, we re-trained our model on five different seeds in total to test its robustness with respect to the particular choice of seed. We report the results in Tab.~\ref{tab:table_1_seeds}. These results indicate that our method is statistically significant with respect to previous SOTA approaches.

\setlength{\tabcolsep}{2pt}
\begin{table*}[!t]
    \caption{We augment Tab.~\ref{tab:gen_sprites_mug_timit} for the MUG dataset with the mean and standard deviation measures using of five models trained with different seed values.}
    \label{tab:table_1_seeds}
    
    \resizebox{\textwidth}{!}{
    \centering
    \footnotesize
        \begin{tabular}[t]{lcccc|cccc}
            \toprule
            & \multicolumn{4}{c|}{Sprites} & \multicolumn{4}{c}{MUG} \\
            Method & Acc$\uparrow$ & IS$\uparrow$ & $H(y|x){\downarrow}$ & $H(y){\uparrow}$ & Acc$\uparrow$ & IS$\uparrow$ & $H(y|x){\downarrow}$ & $H(y){\uparrow}$ \\
            \midrule
            MoCoGAN & $92.89\%$ & $8.461$ & $0.090$ & $2.192$ & $63.12\%$ & $4.332$ & $0.183$ & $1.721$ \\
            DSVAE & $90.73\%$ & $8.384$ & $0.072$ & $2.192$ & $54.29\%$ & $3.608$ & $0.374$ & $1.657$ \\
            R-WAE & $98.98\%$ & $8.516$ & $0.055$ & $\boldsymbol{2.197}$ & $71.25\%$ & $5.149$ & $0.131$ & $1.771$ \\
            S3VAE & $99.49\%$ & $8.637$ & $0.041$ & $\boldsymbol{2.197}$ & $70.51\%$ & $5.136$ & $0.135$ & $1.760$ \\
            SKD & $\boldsymbol{100\%}$ & $\boldsymbol{8.999}$ & $\boldsymbol{\expnumber{1.6}{-7}}$ & $\boldsymbol{2.197}$ & $77.45\%$ & $\boldsymbol{5.569}$ & $\boldsymbol{0.052}$ & $1.769$ \\
            C-DSVAE & $99.99\%$ & $8.871$ & $0.014$ & $\boldsymbol{2.197}$ & $81.16\%$ & $5.341$ & $0.092$ & $1.775$ \\
            \midrule
            Ours & $\boldsymbol{100\% \pm 0\%}$ & $8.942 \pm \num{3.3e-5}$ & $0.006 \pm \num{4e-6}$ & $\boldsymbol{2.197 \pm 0}$ & $\boldsymbol{85.06\% \pm 1.06}$ & $5.517 \pm 0.034$ & $0.073 \pm \num{4e-3}$ & $\boldsymbol{1.782 \pm \num{3e-3}}$ \\
            \bottomrule
        \end{tabular} }
\end{table*}

\end{document}